\documentclass[12pt]{article}

\usepackage{graphicx}
\RequirePackage[OT1]{fontenc}
\RequirePackage{amsthm,amsmath}
\usepackage{indentfirst}

\numberwithin{equation}{section}
\theoremstyle{plain}
\newtheorem{thm}{Theorem}

\newtheorem{lem}{Lemma}

\newcommand{\noi}{\noindent}

\newcommand{\wht}{\widehat}

\newcommand{\wbox}{\sqcap\llap{$\sqcup$}}

\begin{document}

\renewcommand{\baselinestretch}{2}

\fontsize{12}{14pt plus.8pt minus .6pt}\selectfont \vspace{0.8pc}
\centerline{\large\bf INFINITE ARMS BANDIT:}
\vspace{2pt} \centerline{\large\bf OPTIMALITY VIA CONFIDENCE BOUNDS }
\vspace{.4cm} \centerline{Hock Peng Chan and Hu Shouri} \vspace{.4cm} \centerline{\it
National University of Singapore} \vspace{.55cm} \fontsize{9}{11.5pt plus.8pt minus
.6pt}\selectfont





\begin{quotation}
Berry et al. (1997) initiated the development of the infinite arms bandit problem.
They derived a regret lower bound of all allocation strategies for Bernoulli rewards with uniform priors, 
and proposed strategies based on success runs.
Bonald and Prouti\`{e}re (2013) proposed a two-target algorithm that achieves the regret lower bound,  
and extended optimality to Bernoulli rewards with general priors.
We present here a confidence bound target (CBT) algorithm that achieves optimality for rewards that are bounded above. 
For each arm we construct a confidence bound and compare it against each other and a target value to determine if the arm should be sampled further. 
The target value depends on the assumed priors of the arm means. 
In the absence of information on the prior, 
the target value is determined empirically.
Numerical studies here show that CBT is versatile and outperforms its competitors.

\vspace{9pt}
\end{quotation}\par

\fontsize{12}{14pt plus.8pt minus .6pt}\selectfont

\section{Introduction}

Berry, Chen, Zame, Heath and Shepp (1997) initiated the development of the infinite arms bandit problem. 
They showed in the case of Bernoulli rewards with uniform prior a $\sqrt{2n}$ regret lower bound for $n$ rewards,
and provided algorithms based on success runs that achieve no more than $2 \sqrt{n}$ regret.
Bonald and Prouti\`{e}re (2013) provided a two-target stopping-time algorithm that can get arbitrarily close to Berry et al.'s lower bound, 
and is also optimal on Bernoulli rewards with general priors. 
Wang, Audibert and Munos (2008) considered bounded rewards and showed that their confidence bound algorithm has regret bounds that are $\log n$ times the optimal regret. 
Vermorel and Mohri (2005) proposed a POKER algorithm for general reward distributions and priors.

The confidence bound method is arguably the most influential approach for the fixed arm-size multi-armed bandit problem over the past thirty years.
Lai and Robbins (1985) derived the smallest asymptotic regret that a multi-armed bandit algorithm can achieve.
Lai (1987) showed that by constructing an upper confidence bound (UCB) for each arm,
playing the arm with the largest UCB,
this smallest regret is achieved in exponential families. 
The UCB approach was subsequently extended to unknown time-horizons and other parametric families in Agrawal (1995a), 
Auer, Cesa-Bianchi and Fischer (2002),
Burnetas and Katehakis (1996),
Capp\'{e}, Garivier, Maillard, Munos and Stoltz (2013) and Kaufmann, Capp\'{e} and Garivier (2012),
and it has been shown to perform well in practice, 
achieving optimality beyond exponential families.
Chan (2019) modified the subsampling approach of Baransi, Maillard and Mannor (2014) and showed that optimality is achieved in exponential families, 
despite not applying parametric information in the selection of arms. 
The method can be considered to be applying confidence bounds that are computed empirically from subsample information,
which substitutes for the missing parametric information.
A related problem is the study of the multi-armed bandit with irreversible constraints, 
initiated by Hu and Wei (1989).

Good performances and optimality has also been achieved by Bayesian approaches to the multi-armed bandit problem,
see Berry and Fridstedt (1985), Gittins (1989) and Thompson (1933) for early groundwork on the Bayesian approach,
and Korda, Kaufmann and Munos (2013) for more recent advances.
 
In this paper we show how the confidence bound method can be applied on the infinite arms bandit problem. 
We call this new procedure confidence bound target (CBT).
Like the UCB algorithm for finite number of arms,
in CBT a confidence bound is computed for each arm.
What is different in CBT is that we specify an additional target value.
We compare the confidence bound of an arm against the target value to decide whether to play an arm further,
or to discard it and play a new arm.

We derive a universal regret lower bound that applies to all infinite arms bandit algorithms.
We then proceed to show how to choose the target value in CBT so as to achieve the regret lower bound.
This optimal target depends only on the prior distribution of the arm means and not on the reward distributions.
That is the reward distributions need not be specified and optimality is still achieved.

To handle the situation in which the prior is not available,
we provide an empirical version of CBT in which the target value is computed empirically.
Numerical studies on Bernoulli rewards and on a URL dataset show that CBT and empirical CBT outperform their competitors.

In a related continuum-armed bandit problem,
there are uncountably infinite number of arms.
Each arm is indexed by a known parameter $\theta$ and has rewards with mean $f(\theta)$, 
where $f$ is an unknown continuous function.
For solutions to the problem of maximizing the expected sum of rewards,
see Agrawal (1995b), Kleinberg (2005), Auer, Ortner and Szepesvari (2007), Cope (2009) and Tyagi and G\"artner (2013). 

A modern development in the finite arms bandit problem is the study of the contextual bandit,
with covariates present to help in the choice of optimal arms.
For references on this see for example Yang and Zhu (2002), Wang, Kulkarni and Poor (2005), Goldenshluger and Zeevi (2013), Perchet and Rigollet (2013) and Slivkins (2014).

The layout of the paper is as follows.
In Section 2 we provide the set-up for the infinite arms bandit problem.
In Section~3 we review the infinite arms bandit literature.
In Section~4 we describe CBT.
In Section 5 we motivate why the chosen target of CBT leads to the regret lower bound,
and state the optimality of CBT.
In Section 6 we introduce an empirical version of CBT to tackle unknown priors and explain why it works. 
In Section~7 we perform numerical studies.
In Section 8 we summarize a supplementary document to this paper.

\section{Problem set-up}

Let $Y_{k1}, Y_{k2}, \ldots$ be i.i.d. rewards from an arm or population $\Pi_k$.
In the classical multi-armed bandit problem,
there are finitely many arms and the objective is to sequentially select the arms so as to maximize expected sum of rewards.
Equivalently we minimize the regret,
which is the expected cumulative differences between the best arm mean and the mean of the arm played.

In the infinite arms bandit problem that we consider here,
there are infinitely many arms and the rewards are bounded above by 1.
Hence $Y_{kt} \leq 1$ for all $k$ and $t$.
A special example is Bernoulli rewards.

The regret of a bandit algorithm,
after $n$ trials,
is defined to be 
\begin{equation} \label{Rn}
R_n = E \Big( \sum_{k=1}^{\infty} \sum_{t=1}^{n_k} X_{kt} \Big), \mbox{ where } X_{kt} = 1-Y_{kt} (\geq 0)
\end{equation}
is the loss associated with reward $Y_{kt}$,
and $n_k$ is the number of times arm $k$ has been played (hence $n=\sum_{k=1}^{\infty} n_k$).
A key difference between (\ref{Rn}) and the regret of the classical multi-armed bandit problem is that in the classical problem,
the expectation is with respect to fixed distributions,  
whereas in (\ref{Rn}) the expectation is with respect to a Bayesian framework that we shall now describe.

Let $\mu_k$ denote the mean of the loss $X_{kt}$.
The expectation in (\ref{Rn}) is with respect to $\mu_1, \mu_2, \ldots$ i.i.d. with (prior) density $g$ on $(0,\infty)$.
Each $\mu$ for which $g(\mu)>0$ represents a specific distribution $F_{\mu}$ with mean $\mu$.
Hence given $\mu_k$,
$$X_{kt} \stackrel{\rm i.i.d.}{\sim} F_{\mu_k} \mbox{ for } t \geq 1.
$$
We do not assume however that $F_{\mu}$ is known. 
In the infinite arms bandit problem a key decision to be made at each trial is whether to sample a new arm or to play a previously played arm.
The Bayesian framework is helpful in providing some information on the new arms.

In Agrawal and Goyal (2012),
Bubeck and Liu (2013) and Russo and van Roy (2014),
the finite armed bandit problem was studied in which arm means are random,
as described above.
They termed this the {\it stochastic bandit problem} and termed the regret with random arm means {\it Bayesian regret}.

\section{Preliminary background}

Let $a \wedge b$ denote $\min(a,b)$,
$\lfloor \cdot \rfloor$ ($\lceil \cdot \rceil$) denote the greatest (least) integer function and $a^+$ denote $\max(0,a)$.
We say that $a_n \sim b_n$ if $\lim_{n \rightarrow \infty} (a_n/b_n)=1$,
$a_n = o(b_n)$ if $\lim_{n \rightarrow \infty} (a_n/b_n)=0$, 
and $a_n = O(b_n)$ if $\limsup_{n \rightarrow \infty} |a_n/b_n| < \infty$.

Berry et al. (1997) showed that for Bernoulli rewards with $g$ uniform on $(0,1)$,
a regret lower bound 
\begin{equation} \label{liminf}
\liminf_{n \rightarrow \infty} \tfrac{R_n}{\sqrt{n}} \geq \sqrt{2}
\end{equation}
is unavoidable. 
They proposed the following bandit strategies.

\begin{enumerate}
\item $f$-failure strategy.
We play the same arm until $f$ failures are encountered. 
When this happens we switch to a new arm.
We do not go back to a previously played arm,
that is the strategy is {\it non-recalling}. 

\item $s$-run strategy.
We restrict ourselves to no more than $s$ arms,
following the 1-failure strategy in each, 
until a success run of length $s$ is observed in an arm.
When this happens we play the arm for the remaining trials.
If no success run of length $s$ is observed in all $s$ arms,
then the arm with the highest proportion of success is played for the remaining trials.

\item Non-recalling $s$-run strategy.
We follow the 1-failure strategy until an arm produces a success run of length $s$.
When this happens we play the arm for the remaining trials. 
If no arm produces a success run of length $s$, 
then the 1-failure strategy is used for all $n$ trials.

\item $m$-learning strategy.
We follow the 1-failure strategy for the first $m$ trials,
with the arm at trial $m$ played until it yields a failure.
Thereafter we play, 
for the remaining trials, 
the arm with the highest proportion of success. 
\end{enumerate}

Berry et al. showed that $R_n \sim n/(\log n)$ for the $f$-failure strategy for any $f \geq 1$,
whereas for the $\sqrt{n}$-run strategy, 
the $n^{\frac{1}{2}} \log n$-learning strategy and the non-recalling $\sqrt{n}$-run strategy,
$$\limsup_{n \rightarrow \infty} \tfrac{R_n}{\sqrt{n}} \leq 2.
$$

Bonald and Prouti\`{e}re (2013) proposed a two-target algorithm with target values $s_1 = \lfloor \sqrt[3]{\frac{n}{2}} \rfloor$ and $s_f = \lfloor f \sqrt{\frac{n}{2}} \rfloor$,
where $f \geq 2$ is user-defined. 
An arm is discarded if it has fewer than $s_1$ successes when it encounters its first failure,
or $s_f$ successes when it encounters its $f$th failure.
If both targets are met,
then the arm is accepted and played for the remaining trials.
Bonald and Prouti\`{e}re showed that for the uniform prior, 
the two-target algorithm satisfies,
for $n \geq \tfrac{f^2}{2}$,
$$R_n \leq f+(\tfrac{s_f+1}{f}) (\tfrac{s_f-f+2}{s_f-s_1-f+2})^f (2+\tfrac{1}{f}+\tfrac{2(f+1)}{s_1+1}),
$$
from which they conclude that
\begin{equation} \label{lsR}
\limsup_{n \to \infty} \tfrac{R_n}{\sqrt{n}}  \leq \sqrt{2} + \tfrac{1}{f \sqrt{2}}.
\end{equation}
By considering $f$ large,
the regret lower bound in (\ref{liminf}) is achieved.

Bonald and Prouti\`{e}re extended their optimality on Bernoulli rewards to non-uniform priors.
They showed that when $g(\mu) \sim \alpha \mu^{\beta-1}$ for some $\alpha>0$ and $\beta>0$ as $\mu \rightarrow 0$,
the regret lower bound of Berry et al. can be extended to 
\begin{equation} \label{liminf2}
\liminf_{n \rightarrow \infty} (n^{-\frac{\beta}{\beta+1}} R_n) \geq C_0,
\mbox{ where } C_0 = (\tfrac{\beta(\beta+1)}{\alpha})^{\frac{1}{\beta+1}}.
\end{equation}
They also showed that their two-target algorithm with $s_1 = \lfloor n^{\frac{1}{\beta+2}} C_0^{-\frac{\beta+1}{\beta+2}} \rfloor$ 
and $s_f = \lfloor f n^{\frac{1}{\beta+1}} C_0^{-1} \rfloor$ satisfies
\begin{equation} \label{limsup2}
\limsup_{f \rightarrow \infty} [\limsup_{n \rightarrow \infty} (n^{-\frac{\beta}{\beta+1}} R_n)] \leq C_0.
\end{equation}

Wang, Audibert and Munos (2008) proposed a UCB-F algorithm for rewards taking values in $[0,1]$.
They showed that if 
$$P_g(\mu_k \leq \mu) = O(\mu^{\beta}) \mbox{ as } \mu \rightarrow 0 \mbox{ for some } \beta>0,
$$
then under suitable regularity conditions,
$R_n = O(n^{\frac{\beta}{\beta+1}} \log n)$.
In UCB-F an order $n^{\frac{\beta}{\beta+1}}$ arms are chosen, 
and confidence bounds are computed on these arms to determine which arm to play. 
UCB-F is different from CBT in that it pre-selects the number of arms,
and it also does not have a mechanism to reject weak arms quickly. 
Carpentier and Valko (2014) also considered rewards taking values in [0,1],
but their interest in maximizing the selection of a good arm differs from the aims here and in the papers above.

\section{Proposed methodology}

We propose here CBT in which a confidence bound for $\mu_k$ is constructed for each arm $k$, 
and we play an arm as long as its confidence bound is below a specified target.
The confidence bounds are computed using positive confidence coefficients $b_n$ and $c_n$ satisfying 
\begin{equation} \label{bncn}
b_n \rightarrow \infty \mbox{ and } c_n \rightarrow \infty \mbox{ with } b_n+c_n=o(n^{\delta}) \mbox{ for all } \delta>0.
\end{equation}
In our numerical studies we select $b_n = c_n = \log \log n$.

Let $S_{kt} = \sum_{u=1}^t X_{ku}$, 
$\bar X_{kt} = t^{-1} S_{kt}$ and $\wht \sigma^2_{kt} = t^{-1} \sum_{u=1}^t (X_{ku}-\bar X_{kt})^2$.
For an arm $k$ that has been played $t$ times,
its confidence bound is defined to be 
\begin{equation} \label{Lkt}
L_{kt} = \max \Big( \frac{\bar X_{kt}}{b_n}, \bar X_{kt} - c_n \frac{\wht \sigma_{kt}}{\sqrt{t}} \Big).
\end{equation}
Let $\zeta>0$ be a specified target of $L_{kt}$.
In CBT the arms are played sequentially. 
An arm $k$ is played until $L_{kt}$ goes above $\zeta$ and it is discarded when that happens.
We discuss in Section 5 how $\zeta$ should be selected to achieve optimality.
It suffices to mention here that the optimal $\zeta$ decreases at a polynomial rate with respect to $n$. 

\vskip0.9in
\underline{Confidence bound target (CBT)}

\smallskip
\begin{enumerate}
\item Play arm 1 at trial 1.

\item For $m=1,\ldots,n-1$:
Let $k$ be the arm played at trial $m$,
and let $t$ be the number of times arm $k$ has been played up to trial $m$.

\begin{enumerate}
\item If $L_{kt} \leq \zeta$,
then play arm $k$ at trial $m+1$.

\item If $L_{kt} > \zeta$,
then play arm $k+1$ at trial $m+1$.
\end{enumerate}
\end{enumerate}

Let $K$ be the number of arms played after $n$ trials,
and let $n_k$ be the number of times arm $k$ has been played after $n$ trials.
Hence $n=\sum_{k=1}^K n_k$.

\vskip0.5in
There are three types of arms that we need to take care of,
and that explains the design of $L_{kt}$.
The first type are arms with $\mu_k$ (mean of loss $X_{kt}$) significantly larger than $\zeta$.
For these arms we would like to reject them quickly.
The condition that an arm be rejected when $\tfrac{\bar X_{kt}}{b_n}$ exceeds $\zeta$ is key to achieving this.

The second type are arms with $\mu_k$ larger than $\zeta$ but not by as much as those of the first type.
We are unlikely to reject these arms quickly as it is difficult to determine whether $\mu_k$ is smaller or larger than $\zeta$ based on a small sample.
Rejecting arm $k$ when $\bar X_{kt} - c_n \tfrac{\wht \sigma_{kt}}{\sqrt{t}}$ exceeds $\zeta$ 
ensures that arm~$k$ is rejected only when it is statistically significant that $\mu_k$ is larger than $\zeta$.
Though there may be large number of rewards from these arms,
their contributions to the regret are small because these arms have small $\mu_k$,
as $\zeta$ is chosen small when $n$ is large. 

The third type of arms are those with $\mu_k$ smaller than $\zeta$.
For these arms the best strategy (when $\zeta$ is chosen correctly) is to play them for the remaining trials.
Selecting $b_n \rightarrow \infty$ and $c_n \rightarrow \infty$ in (\ref{Lkt}) ensures that the probabilities of rejecting these arms are small.

For Bernoulli rewards the first target $s_1$ of the two-target algorithm is designed for quick rejection of the first type of arms, 
and the second target $s_f$ is designed for rejection of the second type.
What is different is that whereas two-target monitors an arm for rejection only when there are 1 and $f$ failures,
with $f$ chosen large for optimality,
CBT checks for rejection each time a failure occurs.
The frequent monitoring of CBT is a positive feature that results in significantly better performances in the numerical experiments of Section 7. 

\section{Optimality}

We state the regret lower bound in Section 5.1 and show that CBT achieves this bound in Section 5.2.

\subsection{Regret lower bound}

In Lemma \ref{lem1} below we motivate the choice of $\zeta$.
Let $P_{\mu}$ denote probability and $E_{\mu}$ denote expectation,
with respect to $X \stackrel{d}{\sim} F_{\mu}$.
Let $P_g(\cdot) = \int_0^{\infty} P_{\mu}(\cdot) g(\mu) d \mu$ and $E_g(\cdot) = \int_0^{\infty} E_{\mu}(\cdot) g(\mu) d \mu$.
Let $\lambda = \int_0^{\infty} E_{\mu}(X|X>0) g(\mu) d \mu [=E_g(X|X>0)]$ be the mean of the first positive loss of a random arm.
We assume that $\lambda<\infty$.
The value $\lambda$ is the unavoidable cost of exploring a new arm.
We consider $E_{\mu}(X|X>0)$ instead of $\mu$ because it makes sense to reject an arm only after observing a positive loss.
For Bernoulli rewards $\lambda=1$.
Let $p(\zeta) = P_g(\mu_1 \leq \zeta)$ and $v(\zeta) = E_g(\zeta-\mu_1)^+$.

Consider an idealized algorithm which plays $\Pi_k$ until a positive loss is observed,
and $\mu_k$ is revealed when that happens. 
If $\mu_k > \zeta$,
then $\Pi_k$ is rejected and a new arm is played next. 
If $\mu_k  \leq \zeta$,
then we stop exploring and play $\Pi_k$ for the remaining trials.

Let
\begin{equation} \label{hmus}
r_n(\zeta) = \tfrac{\lambda}{p(\zeta)} + n E_g(\mu_1|\mu_1 \leq \zeta).
\end{equation}
Assuming that the exploration stage uses $o(n)$ trials and $\zeta$ is small,
the regret of this algorithm is asymptotically $r_n(\zeta)$.
Let $K$ be the total number of arms played.
The first term in the expansion of $r_n(\zeta)$ approximates $E(\sum_{k=1}^{K-1} \sum_{t=1}^{n_k} X_{kt})$ whereas the second term approximates 
$E(\sum_{t=1}^{n_K} X_{Kt})$.

\begin{lem} \label{lem1}
Let $\zeta_n$ be such that $v(\zeta_n) = \frac{\lambda}{n}$. 
We have
$$\inf_{\zeta > 0} r_n(\zeta) = r_n(\zeta_n) = n \zeta_n.
$$
\end{lem}

\smallskip
{\sc Proof}. 
Since $E_g(\zeta-\mu_1|\mu_1 \leq \zeta) = \tfrac{v(\zeta)}{p(\zeta)}$,
it follows from (\ref{hmus}) that
\begin{equation} \label{h2}
r_n(\zeta) = \tfrac{\lambda}{p(\zeta)}+n \zeta-\tfrac{nv(\zeta)}{p(\zeta)}.
\end{equation}
It follows from $\tfrac{d}{d \zeta} v(\zeta)= p(\zeta)$ and $\tfrac{d}{d \zeta} p(\zeta) = g(\zeta)$ that
$$\tfrac{d}{d \zeta} r_n(\zeta) = \tfrac{g(\zeta)[nv(\zeta)-\lambda]}{p^2(\zeta)}.
$$
Since $v$ is continuous and strictly increasing when it is positive,
the root to $v(\zeta)=\frac{\lambda}{n}$ exists,
and Lemma \ref{lem1} follows from solving $\tfrac{d}{d \zeta} r_n(\zeta)=0$.
$\wbox$

\smallskip
Consider: 

\smallskip
\noindent (A1) There exists $\alpha > 0$ and $\beta>0$ such that $g(\mu) \sim \alpha \mu^{\beta-1}$ as $\mu \to 0$.

\smallskip 
\noindent Under (A1),
$p(\zeta) = \int_0^{\zeta} g(\mu) d \mu \sim \frac{\alpha}{\beta} \zeta^{\beta}$ and $v(\zeta) = \int_0^{\zeta} p(\mu) d \mu \sim \frac{\alpha}{\beta (\beta+1)} \zeta^{\beta+1}$ 
as $\zeta \rightarrow 0$,
hence $v(\zeta_n) \sim \tfrac{\lambda}{n}$ for  
\begin{equation} \label{mustar}
\zeta_n \sim C n^{-\frac{1}{\beta+1}}, \mbox{ where } C = (\tfrac{\lambda \beta(\beta+1)}{\alpha})^{\frac{1}{\beta+1}}.
\end{equation}

\smallskip
In Lemma \ref{thm1} below we state the regret lower bound.
We assume there that:

\smallskip
\noindent (A2) There exists $a_1>0$ such that $P_{\mu}(X>0) \geq a_1 \min(\mu,1)$ for all $\mu$.

\smallskip
\noindent With (A2) we avoid bad arms which are played a large number of times because their losses are mostly zeros but are very big when positive.

\begin{lem} \label{thm1}
Under {\rm (A1)} and {\rm (A2)},
all infinite arms bandit algorithms have regret satisfying
\begin{equation} \label{R2}
R_n \geq [1+o(1)] n \zeta_n  (\sim Cn^{\tfrac{\beta}{\beta+1}}) \mbox{ as } n \rightarrow \infty.   
\end{equation}
\end{lem}

\smallskip
{\sc Example} 1.
Consider $X \stackrel{d}{\sim}$ Bernoulli($\mu$).
Condition (A2) holds with $a_1=1$.
If $g$ is uniform on (0,1),
then (A1) holds with $\alpha=\beta=1$. 
Since $\lambda=1$,
by (\ref{mustar}),
$\zeta_n \sim \sqrt{\tfrac{2}{n}}$.
Lemma \ref{thm1} says that $R_n \geq [1+o(1)] \sqrt{2n}$,
agreeing with Theorem 3 of Berry et al. (1997).
\smallskip

Bonald and Prouti\`{e}re (2013) showed (\ref{R2}) in their Lemma 3 for Bernoulli rewards under (A1) and that their two-target algorithm gets close to the regret lower bound when $f$ is large.
It will be shown in Theorem \ref{thm2} that the lower bound in (\ref{R2}) is achieved by CBT for rewards that need not be Bernoulli.

\subsection{Optimality of CBT}

We state the optimality of CBT in Theorem \ref{thm2},
after describing the conditions on discrete rewards under (B1) and continuous rewards under (B2) for which the theorem holds.
Let $M_{\mu}(\theta) = E_{\mu} e^{\theta X}$.

\smallskip
\noindent (B1) The rewards are integer-valued.
For $0 < \delta \leq 1$,
there exists $\theta_{\delta}>0$ such that for $\mu>0$ and $0 \leq \theta \leq \theta_{\delta}$,
\begin{eqnarray} \label{Mbound}
M_{\mu}(\theta) & \leq & e^{(1+\delta) \theta \mu}, \\ \label{Mmu2}
M_{\mu}(-\theta) & \leq & e^{-(1-\delta) \theta \mu}.
\end{eqnarray}
In addition,
\begin{eqnarray} \label{a2}
P_{\mu}(X>0) & \leq & a_2 \mu \mbox{ for some } a_2>0, \\ \label{E4}
E_{\mu} X^4 & = & O(\mu) \mbox{ as } \mu \rightarrow 0.
\end{eqnarray}

\smallskip
\noindent (B2) The rewards are continuous random variables satisfying
\begin{equation} \label{Prho}
\sup_{\mu>0} P_{\mu}(X \leq \gamma \mu) \rightarrow 0 \mbox{ as } \gamma \rightarrow 0.
\end{equation}
Moreover (\ref{E4}) holds and for $0 < \delta \leq 1$,
there exists $\tau_{\delta}>0$ such that for $0 < \theta \mu \leq \tau_{\delta}$,
\begin{eqnarray} \label{tau1}
M_{\mu}(\theta) & \leq & e^{(1+\delta) \theta \mu}, \\ \label{tau2}
M_{\mu}(-\theta) & \leq & e^{-(1-\delta) \theta \mu}. 
\end{eqnarray}
In addition for each $t \geq 1$, 
there exists $\xi_t >0$ such that
\begin{equation} \label{sigbound}
\sup_{\mu \leq \xi_t} P_{\mu}(\wht \sigma_t^2 \leq \gamma \mu^2) \rightarrow 0 \mbox{ as } \gamma \rightarrow 0,
\end{equation}
where $\wht \sigma_t^2 = t^{-1} \sum_{u=1}^t (X_u -\bar X_t)^2$ and $\bar X_t = t^{-1} \sum_{u=1}^t X_u$ for i.i.d. $X_u \stackrel{d}{\sim} F_{\mu}$.

\begin{thm} \label{thm2}
Assume {\rm (A1), (A2)} and either {\rm (B1)} or {\rm (B2)}. 
For CBT with threshold $\zeta_n$ satisfying {\rm (\ref{mustar})} and $b_n$,
$c_n$ satisfying {\rm (\ref{bncn})}, 
\begin{equation} \label{R1}
R_n \sim n \zeta_n \mbox{ as } n \rightarrow \infty.  
\end{equation}
\end{thm}

Theorem \ref{thm2} says that CBT is optimal as it attains the lower bound given in Lemma \ref{thm1}.
In the examples below we show that the regularity conditions (A2),
(B1) and (B2) are reasonable and checkable.

\smallskip
{\sc Example} 2.
If $X \stackrel{d}{\sim}$ Bernoulli($\mu$) under $P_{\mu}$,
then 
$$M_{\mu}(\theta) = 1-\mu+\mu e^{\theta} \leq \exp[\mu(e^{\theta}-1)].
$$
Hence (\ref{Mbound}) and (\ref{Mmu2}) hold with $\theta_{\delta} >0$ satisfying
\begin{equation} \label{etheta}
e^{\theta_{\delta}}-1 \leq \theta_{\delta}(1+\delta) \mbox{ and } e^{-\theta_{\delta}}-1 \leq -\theta_{\delta}(1-\delta).
\end{equation}
In addition (\ref{a2}) holds with $a_2=1$,
and (\ref{E4}) holds because $E_{\mu}X^4 = \mu$.
Condition (A2) holds with $a_1=1$.

\smallskip
{\sc Example} 3.
Let $F_{\mu}$ be a distribution with support on $0, \ldots, I$ for some positive integer $I>1$ and having mean $\mu$.
Let $p_i = P_{\mu}(X=i)$.
We check that $P_{\mu}(X>0) \geq \mu I^{-1}$ and therefore (A2) holds with $a_1=I^{-1}$.

Let $\theta_{\delta}>0$ be such that
\begin{equation} \label{Id}
e^{i \theta}-1 \leq i \theta(1+\delta) \mbox{ and } e^{-i \theta}-1 \leq -i \theta(1-\delta) \mbox{ for } 0 \leq i \theta \leq I \theta_{\delta}.
\end{equation}
By (\ref{Id}) for $0 \leq \theta \leq \theta_{\delta}$,
\begin{eqnarray*}
M_{\mu}(\theta) & = & \sum_{i=0}^I p_i e^{i \theta} \leq 1+(1+\delta) \mu \theta, \cr
M_{\mu}(-\theta) & = & \sum_{i=0}^I p_i e^{-i \theta} \leq 1-(1-\delta) \mu \theta, \cr
\end{eqnarray*}
and (\ref{Mbound}), 
(\ref{Mmu2}) follow from $1+x \leq e^x$.
Moreover (\ref{a2}) holds with $a_2=1$ and (\ref{E4}) holds because $E_{\mu} X^4 = \sum_{i=0}^I p_i i^4 \leq I^3 \mu$.

\smallskip
{\sc Example} 4.
If $X \stackrel{d}{\sim}$ Poisson($\mu$),
then  
$$M_{\mu}(\theta) = \exp[\mu(e^{\theta}-1)],
$$
and (\ref{Mbound}), (\ref{Mmu2}) again follow from (\ref{etheta}).
Since $P_{\mu}(X>0) = 1-e^{-\mu}$,
(A2) holds with $a_1=1-e^{-1}$,
and (\ref{a2}) holds with $a_2=1$.
In addition (\ref{E4}) holds because
$$E_{\mu} X^4 = \sum_{k=1}^{\infty} \tfrac{k^4 \mu^k e^{-\mu}}{k!} = \mu e^{-\mu} + e^{-\mu} O \Big( \sum_{k=2}^{\infty} \mu^k \Big).
$$

{\sc Example} 5.
Let $Z$ be a continuous non-negative random variable with mean 1,
and with $Ee^{\tau_0 Z} < \infty$ for some $\tau_0 > 0$.
Consider $X$ distributed as $\mu  Z$.
Condition (A2) holds with $a_1=1$.
We conclude (\ref{Prho}) from  
$$\sup_{\mu>0} P_{\mu}(X \leq \gamma \mu) = P(Z \leq \gamma) \rightarrow 0 \mbox{ as } \gamma \rightarrow 0.
$$
Let $0 < \delta \leq 1$.
Since $\lim_{\tau \rightarrow 0} \tau^{-1} \log Ee^{\tau Z} = EZ = 1$,
there exists $\tau_{\delta}>0$ such that for $0 < \tau \leq \tau_{\delta}$,
\begin{equation} \label{EetY}
E e^{\tau Z} \leq e^{(1+\delta) \tau} \mbox{ and } E e^{-\tau Z} \leq e^{-(1-\delta) \tau}.
\end{equation}
Since $M_{\mu}(\theta) = E_{\mu} e^{\theta X} = E e^{\theta \mu Z}$ and $M_{\mu}(-\theta) = Ee^{-\theta \mu Z}$,
we conclude (\ref{tau1}) and (\ref{tau2}) from (\ref{EetY}) with $\tau=\theta \mu$. 
We conclude (\ref{E4}) from $E_{\mu} X^4 = \mu^4 EZ^4$,
and (\ref{sigbound}),
for arbitrary $\xi_t>0$,
from
$$P_{\mu}(\wht \sigma_t^2 \leq \gamma \mu^2) = P(\wht \sigma_{tZ}^2 \leq \gamma) \rightarrow 0 \mbox{ as } \gamma \rightarrow 0,
$$
where $\wht \sigma^2_{tZ} = t^{-1} \sum_{u=1}^t (Z_u - \bar Z_t)^2$,
for i.i.d. $Z$ and $Z_u$.  

\section{Methodology for unknown priors}

The optimal implementation of CBT,
in particular the computation of the optimal target $\zeta_n$,
assumes knowledge of how $g(\mu)$ behaves for $\mu$ near 0.
For $g$ unknown we rely on Theorem~\ref{thm2} to motivate the empirical implementation of CBT.

What is striking about (\ref{R1}) is that it relates the optimal target $\zeta_n$ to $\frac{R_n}{n}$, 
and moreover this relation does not depend on either the prior $g$ or the reward distributions.
We suggest therefore,
in an empirical implementation of CBT,
to apply targets
\begin{equation} \label{hmustar}
\zeta(m):= \tfrac{S_m'}{n},
\end{equation}
where $S_m'$ is the sum of the losses $X_{kt}$ over the first $m$ trials.

In the beginning with $m$ small,
$\zeta(m)$ underestimates the optimal target,
but this will only encourage exploration, 
which is the right strategy at the beginning.
As $m$ increases $\zeta(m)$ gets closer to the optimal target,
and empirical CBT behaves more like CBT in deciding whether to play an arm further.
Empirical CBT decides from among all arms played which to play further,
unlike CBT in which arms are played sequentially.

\vskip0.6in
\underline{Empirical CBT}

\smallskip
Notation: 
When there are $m$ total rewards,
let $n_k(m)$ denote the number of rewards from arm $k$ and let $K_m$ denote the number of arms played.

\smallskip
For $m=0$, 
play arm 1.
Hence $K_1=1$, $n_1(1)=1$ and $n_k(1) = 0$ for $k>1$.

\smallskip
For $m=1,\ldots,n-1$:
\begin{enumerate}
\item If $\min_{1 \leq k \leq K_m} L_{kn_k(m)} \leq \zeta(m)$,
then play the arm $k$ minimizing $L_{kn_k(m)}$.

\item If $\min_{1 \leq k \leq K_m} L_{kn_k(m)} > \zeta(m$),
then play a new arm $K_m+1$.
\end{enumerate}

\vskip0.5in
Empirical CBT,
unlike CBT,
does not achieve the smallest regret.
This is because when a good arm (that is an arm with $\mu_k$ below optimal target) appears early,
we are not sure whether this is due to good fortune or that the prior is disposed towards arms with small $\mu_k$, 
so we explore more arms before we are certain and play the good arm for the remaining trials.
Similarly when no good arm appears after some time,
we may conclude that the prior is disposed towards arms with large $\mu_k$, 
and play an arm with $\mu_k$ above the optimal target for the remaining trials, 
even though it is advantageous to explore further.

As the analysis of the regret of empirical CBT is complicated,
we consider an idealized version of empirical CBT in the supplementary document and derive its asymptotic regret,
to give us a sense of the additional regret when applying CBT empirically.

In the idealized version of empirical CBT,
$\mu_k$ is revealed after the first positive loss of arm $k$ is observed.
The number of arms played is the smallest $K$ satisfying
$$\min_{1 \leq k \leq K} \mu_k \leq \tfrac{K \lambda}{n},
$$
and exploitation of the best arm begins after $\mu_1, \ldots, \mu_K$ have been revealed.
The idealized empirical CBT is like the idealized algorithm described in the beginning of Section 5.1,
but with a target $\zeta = \tfrac{k \lambda}{n}$, 
after $k$ arms have been played. 
This is because $\lambda$ is the mean of the first positive loss of each arm,
so after $k$ arms have been played the sum of losses has mean $k \lambda$.
The idealized empirical CBT is a simplification of CBT that captures the additional regret of empirical CBT over CBT when applying a target that does not depend on the prior.

\begin{thm} \label{thm3}
The idealized empirical CBT has regret
\begin{equation} \label{thm3.1}
R_n' \sim I_{\beta} n \zeta_n,
\end{equation}
where $I_{\beta} = (\frac{1}{\beta+1})^{\frac{1}{\beta+1}} (2-\frac{1}{(\beta+1)^2}) \Gamma(2-\frac{1}{\beta+1})$ and $\Gamma(u) = \int_0^{\infty} x^{u-1} e^{-x} dx$.
\end{thm}

The constant $I_{\beta}$ increases from 1 (at $\beta=0$) to 2 (at $\beta=\infty$).
The increase is quite slow so for reasonable values of $\beta$ it is closer to 1 than 2.
For example $I_1=1.10$, $I_2=1.17$, $I_3=1.24$ and $I_{10}=1.53$.
The predictions from (\ref{thm3.1}), 
that the inflation of the regret increases with $\beta$,
and that it should not be more than 25\% for $\beta=$1, 2 and 3, 
are consistent with the simulation outcomes in Section 7.

\section{Numerical studies}

We study here arms with rewards that have Bernoulli (Example 6) as well as unknown distributions (Example 7).
In our simulations 10,000 datasets are generated for each entry in Tables 1--4,
and standard errors are placed after the $\pm$ sign.
In both CBT and empirical CBT, 
we select $b_n = c_n = \log \log n$.

\begin{table}[h!]
\centering
\scalebox{0.85}{%
\begin{tabular}{ll|rrrr}
\multicolumn{2}{c|}{Algorithm} & \multicolumn{4}{c}{Regret} \cr
& & $n=$100 & $n=$1000 & $n=$10,000 & $n=$100,000 \cr \hline
CBT & $\zeta=\sqrt{2/n}$ & 14.6$\pm$0.1 & 51.5$\pm$0.3 & 162$\pm$1 & 504$\pm$3 \cr
& empirical & 15.6$\pm$0.1 & 54.0$\pm$0.3 & 172$\pm$1 & 531$\pm$3 \cr \hline
Berry et al. & 1-failure & 21.8$\pm$0.1 & 152.0$\pm$0.6 & 1123$\pm$4 & 8955$\pm$28 \cr
& $\sqrt{n}$-run & 19.1$\pm$0.2& 74.7$\pm$0.7 & 260$\pm$3 & 844$\pm$9 \cr
& $\sqrt{n}$-run (non-recall) & 15.4$\pm$0.1 & 57.7$\pm$0.4 & 193$\pm$1 & 618$\pm$4 \cr
& $n^{\frac{1}{2}} \log n$-learning & 18.7$\pm$0.1 & 84.4$\pm$0.6 & 311$\pm$3 & 1060$\pm$9 \cr \hline
Two-target & $f=3$ & 15.2$\pm$0.1 & 52.7$\pm$0.3 & 167$\pm$1 & 534$\pm$3 \cr
& $f=6$ & 16.3$\pm$0.1 & 55.8$\pm$0.4 & 165$\pm$1 & 511$\pm$3 \cr
& $f=9$ & 17.5$\pm$0.1 & 58.8$\pm$0.4 & 173$\pm$1 & 514$\pm$3 \cr \hline
UCB-F & $K= \lfloor \sqrt{n/2} \rfloor$ & 39.2$\pm$0.1 & 206.4$\pm$0.4 & 1204$\pm$1 & 4432$\pm$15 \cr \hline
Lower bound & $\sqrt{2n}$ & 14.1 \quad \ \ \ & 44.7 \quad \ \ \ & 141 \quad \quad & 447 \quad \quad
\end{tabular} }
\caption{The regrets for Bernoulli rewards with uniform prior $(\beta=1)$.}
\end{table}

\begin{table}[h!]
\centering
\scalebox{0.85}{%
\begin{tabular}{ll|rrrr}
& & \multicolumn{4}{c}{Regret} \cr
\multicolumn{2}{c|}{Algorithm} & $n=100$ & $n=1000$ & $n=$10,000 & $n=$100,000 \cr \hline
CBT & $\zeta = Cn^{-\frac{1}{3}}$ & 24.9$\pm$0.1 & 124.8$\pm$0.5 & 575$\pm$3 & 2567$\pm$12 \cr
& empirical & 25.6$\pm$0.1 & 132.3$\pm$0.6 & 604$\pm$2 & 2816$\pm$11 \cr \hline
Two-target & $f=3$ & 25.0$\pm$0.1 & 132.1$\pm$0.6 & 649$\pm$3 & 3099$\pm$16 \cr
& $f=6$ & 26.0$\pm$0.1 & 131.1$\pm$0.6 & 600$\pm$3 & 2783$\pm$13 \cr
& $f=9$ & 26.7$\pm$0.1 & 136.6$\pm$0.7 & 605$\pm$3 & 2676$\pm$12 \cr \hline
UCB-F & & 43.6$\pm$0.1 & 386.8$\pm$0.3 & 2917$\pm$2 & 16038$\pm$12 \cr
$n^{\frac{1}{\beta+1}}$-run & non-recall & 28.1$\pm$0.1 & 172.5$\pm$0.9 & 903$\pm$5 & 4434$\pm$28 \cr \hline
Lower bound & $Cn^{\frac{2}{3}}$ & 23.0 \quad \ \ \ & 106.7 \quad \ \ \ & 495 \quad \quad & 2300 \quad \quad \cr

\end{tabular} }
\caption{The regrets for Bernoulli rewards with $g(\mu) = \frac{\pi}{2} \sin(\pi \mu)$ $(\beta=2)$.}
\end{table}

\begin{table}[h!]
\centering
\scalebox{0.85}{%
\begin{tabular}{ll|rrrr}
& & \multicolumn{4}{c}{Regret} \cr
\multicolumn{2}{c|}{Algorithm} & $n=100$ & $n=1000$ & $n=$10,000 & $n=$100,000 \cr \hline
CBT & $\zeta = Cn^{-\frac{1}{4}}$ & 43.3$\pm$0.1 & 254.8$\pm$0.8 & 1402$\pm$5 & \ 7658$\pm$28 \cr
& empirical & 43.1$\pm$0.1 & 263.8$\pm$0.8 & 1542$\pm$5 & \ 8860$\pm$28 \cr \hline
Two-target & $f=3$ & 43.2$\pm$0.1 & 276.0$\pm$1.0 & 1697$\pm$7 & 10235$\pm$44 \cr
& $f=6$ & 44.5$\pm$0.1 & 270.1$\pm$1.0 & 1537$\pm$6 & 8828$\pm$34 \cr
& $f=9$ & 45.6$\pm$0.1 & 278.5$\pm$1.1 & 1510$\pm$6 & 8501$\pm$33 \cr \hline
UCB-F & & 63.2$\pm$0.1 & 592.9$\pm$0.3 & 5120$\pm$3 & 34168$\pm$25 \cr 
$n^{\frac{1}{\beta+1}}$-run & non-recall & 45.5$\pm$0.2 & 338.2$\pm$1.4 & 2206$\pm$10 & 14697$\pm$73 \cr \hline
Lower bound & $Cn^{\frac{3}{4}}$ & 39.5 \quad \ \ \ & 222.1 \quad \ \ \ & 1249 \quad \quad  & 7022 \quad \quad \cr
\end{tabular} }
\caption{The regrets for Bernoulli rewards with $g(\mu) = 1-\cos(\pi \mu)$ $(\beta=3)$.}
\end{table}

\smallskip
{\sc Example} 6.
We consider Bernoulli rewards with the following priors:

\begin{enumerate}
\item  $g(\mu)=1$,
which satisfies (A1) with $\alpha=\beta=1$,

\item $g(\mu) = \tfrac{\pi}{2} \sin(\pi \mu)$,
which satisfies (A1) with $\alpha=\frac{\pi^2}{2}$ and $\beta=2$,

\item $g(\mu) = 1-\cos(\pi \mu)$,
which satisfies (A1) with $\alpha=\frac{\pi^2}{2}$ and $\beta=3$.
\end{enumerate}

For all three priors,
the two-target algorithm does better with $f=3$ for smaller $n$,
and with $f=6$ or 9 at larger $n$.
CBT is the best performer uniformly over $n$,
and empirical CBT is also competitive against two-target with $f$ fixed.

Even though optimal CBT outperforms empirical CBT,
optimal CBT assumes knowledge of the prior to select the threshold $\zeta$,
which differs with the priors.
On the other hand the same algorithm is used for all three priors when applying empirical CBT,
and in fact the same algorithm is also used on the URL dataset in Example 7,
with no knowledge of the reward distributions.
Hence though empirical CBT is numerically comparable to two-target and weaker than CBT,
it is more desirable as we do not need to know the prior to use it.

For the uniform prior, 
the best performing among the algorithms in Berry et al. (1997) is the non-recalling $\sqrt{n}$-run algorithm.
For UCB-F [cf. Wang et al. (2008)],
the selection of $K = \lfloor (\tfrac{\beta}{\alpha})^{\frac{1}{\beta+1}} (\tfrac{n}{\beta+1})^{\frac{\beta}{\beta+1}} \rfloor$ ($\sim \tfrac{1}{p(\zeta_n)}$)
and ``exploration sequence" ${\cal E}_m = \sqrt{\log m}$ works well.

\begin{table}[h!]
\centering
\scalebox{0.85}{%
\begin{tabular}{ll|ll}
\multicolumn{2}{c|}{ } & \multicolumn{2}{c}{Regret} \cr
Algorithm & $\epsilon$ & $n=$130 & $n=$1300 \cr \hline
emp. CBT & & 212$\pm$2 & 123.8$\pm$0.6 \cr
POKER & & 203 & 132 \cr 
$\epsilon$-greedy & 0.05 & 733 & 431 \cr
$\epsilon$-first & 0.15 & 725 & 411 \cr
$\epsilon$-decreasing & 1.0 & 738 & 411 
\end{tabular} }
\caption{The average regret $\tfrac{R_n}{n}$.}
\end{table}

\smallskip
{\sc Example} 7. We consider here the URL dataset studied in Vermorel and Mohri~(2005),
where a POKER algorithm for dealing with large number of arms is proposed.
We reproduce part of their Table 1 in our Table 4, 
together with new simulations on empirical CBT.
The dataset consists of the retrieval latency of 760 university home-pages,
in milliseconds, with a sample size of more than 1300 for each home-page.
The numbers in the dataset correspond to the non-negative losses $X_{kt}$.
The dataset can be downloaded from ``sourceforge.net/projects/bandit''.

In our simulations the losses for each home-page are randomly permuted in each run.
We see from Table 4 that POKER does better than empirical CBT at $n=130$, 
whereas empirical CBT does better at $n=1300$ .
The other algorithms are uniformly worse than both POKER and empirical CBT.

The algorithm $\epsilon$-first refers to exploring the first $\epsilon n$ losses,
with random selection of the arms to be played.
This is followed by pure exploitation for the remaining $(1-\epsilon) n$ losses, 
on the ``best" arm (with the smallest mean loss).
The algorithm $\epsilon$-greedy refers to selecting,
in each play,
a random arm with probability $\epsilon$,
and the best arm with the remaining $1-\epsilon$ probability. 
The algorithm $\epsilon$-decreasing is like $\epsilon$-greedy except that in the $m$th play,
we select a random arm with probability $\min(1,\frac{\epsilon}{m})$,
and the best arm otherwise.
Both $\epsilon$-greedy and $\epsilon$-decreasing are disadvantaged by not making use of information on the total number of trials.
Vermorel and Mohri also ran simulations on more complicated strategies like LeastTaken, SoftMax, Exp3, GaussMatch and IntEstim,
with average regret ranging from 287--447 for $n=130$ and 189--599 for $n=1300$.

\section{Supplementary Materials}

We provide a supplementary document to support this manuscript. 
It contains the proofs of Lemma \ref{thm1} and Theorems \ref{thm2} and \ref{thm3}.


\begin{thebibliography}{100}

\bibitem{Agr95}
\textsc{Agrawal, R.} (1995a).
Sample mean based index policies with $O(\log n)$ regret for the multi-armed bandit problem.
\textit{Adv. Appl. Probab.} \textbf{17} 1054--1078.

\bibitem{Agr95a}
\textsc{Agrawal, R.} (1995b). 
The continuum-armed bandit problem. 
\textit{SIAM journal on control and optimization} \textbf{33(6)} 1926--1951.

\bibitem{AG12}
\textsc{Agrawal, S.} and  \textsc{Goyal, N.} (2012). 
Analysis of Thompson sampling for the multi-armed bandit problem. 
\textit{In Conference on learning theory} \textbf{23} 39.1--39.26.

\bibitem{ACF02}
\textsc{Auer, P., Cesa-Bianchi, N.} and \textsc{Fischer, P.} (2002).
Finite-time analysis of the multiarmed bandit problem.
\textit{Machine Learning} \textbf{47} 235--256.

\bibitem{AOS07}
\textsc{Auer, P., Ortner, R.,} and \textsc{Szepesv\'ari, C.} (2007). 
Improved rates for the stochastic continuum-armed bandit problem. 
\textit{In International Conference on Computational Learning Theory}  454--468. 

\bibitem{BMM14}
\textsc{Baransi, A., Maillard, O.A.} and \textsc{Mannor, S.} (2014).
Sub-sampling for multi-armed bandits.
\textit{Proceedings of the European Conference on Machine Learning} pp.13.

\bibitem{Berry1997} \textsc{Berry, D., Chen, R., Zame, A., Heath, D.} and \textsc{Shepp, L.} (1997). 
Bandit problems with infinitely many arms. 
\textit{Ann. Statist.}, \textbf{25}, 2103–2116.

\bibitem{BF85}
\textsc{Berry, D.} and \textsc{Fristedt, B.} (1985).
\textit{Bandit problems: sequential allocation of experiments}. 
Chapman and Hall, London. 

\bibitem{Bonald2013} \textsc{Bonald, T.} and \textsc{Prouti\`ere, A.} (2013).
Two-target algorithms for infinite-armed bandits with Bernoulli rewards. 
\textit{Neural Information Processing Systems}.
	
\bibitem{BL02}
\textsc{Brezzi, M.} and \textsc{Lai, T.L.} (2002).
Optimal learning and experimentation in bandit problems.
\textit{J. Econ. Dynamics Cont.} \textbf{27} 87--108.

\bibitem{BL13} 
\textsc{Bubeck, S.} and \textsc{Liu, C. Y.} (2013). 
Prior-free and prior-dependent regret bounds for thompson sampling. 
\textit{In Advances in Neural Information Processing Systems} 638--646.

\bibitem{BK96}
\textsc{Burnetas, A.} and \textsc{Katehakis, M.} (1996).
Optimal adaptive policies for sequential allocation problems.
\textit{Adv. Appl. Math.} \textbf{17} 122--142.


\bibitem{CGMMS13}
\textsc{Capp\'{e}, O., Garivier, A., Maillard, J., Munos, R., Stoltz, G.} (2013).
Kullback-Leibler upper confidence bounds for optimal sequential allocation.
\textit{Ann. Statist.} \textbf{41} 1516--1541.

\bibitem{CV15}
\textsc{Carpentier, A.} and \textsc{Valko, M.} (2015).
Simple regret for infinitely many bandits.
\textit{32th International Conference on Machine Learning}.

\bibitem{Chan18}
\textsc{Chan, H.} (2019).
The multi-armed bandit problem: an efficient non-parametric solution, to appear in {\it Ann. Statist.}

\bibitem{Cope09}
\textsc{Cope, E. W.} (2009). 
Regret and convergence bounds for a class of continuum-armed bandit problems. 
\textit{IEEE Transactions on Automatic Control} \textbf{54(6)} 1243--1253.

\bibitem{Git89}
\textsc{Gittins, J.} (1989).
{\it Multi-armed Bandit Allocation Indices}. 
Wiley, New York.

\bibitem{GZ13}
\textsc{Goldenshluger, A.} and \textsc{Zeevi, A.} (2013). 
A linear response bandit problem. 
\textit{Stochastic Systems} \textbf{3(1)} 230--261.

\bibitem{HW89}
\textsc{Hu, I.} and \textsc{Wei, C.Z.} (1989).
Irreversible adaptive allocation rules. \textit{Ann. Statist.} \textbf{17} 801--822.

\bibitem{KCG12}
\textsc{Kaufmann, E., Capp\'{e}} and \textsc{Garivier, A.} (2012).
On Bayesian upper confidence bounds for bandit problems.
\textit{Proceedings of the Fifteenth International Conference on Artificial Intelligence and Statistics}
\textbf{22} 592--600.

\bibitem{Kle05}
\textsc{Kleinberg, R.D. } (2005).
Nearly tight bounds for the continuum-armed bandit problem. 
\textit{In Advances in Neural Information Processing Systems} 697--704.

\bibitem{KKM13}
\textsc{Korda, N., Kaufmann, E.} and \textsc{Munos, R.} (2013).
Thompson sampling for 1-dimensional exponential family bandits.
\textit{NIPS} \textbf{26} 1448--1456.

\bibitem{Lai87}
\textsc{Lai, T.L.} (1987).
Adaptive treatment allocation and the multi-armed bandit problem.
\textit{Ann. Statist.} \textbf{15} 1091--1114.

\bibitem{LR85}
\textsc{Lai, T.L.} and \textsc{Robbins, H.} (1985).
Asymptotically efficient adaptive allocation rules.
\textit{Adv. Appl. Math.} \textbf{6} 4--22.

\bibitem{PR13} 
\textsc{Perchet, V.} and  \textsc{Rigollet, P.} (2013).
The multi-armed bandit problem with covariates. 
\textit{The Annals of Statistics}  \textbf{41(2)} 693--721.

\bibitem{RR14}	
\textsc{Russo, D.} and \textsc{Van Roy, B.} (2014). 
Learning to optimize via posterior sampling. 
\textit{Mathematics of Operations Research} \textbf{39(4)} 1221--1243.

\bibitem{Sliv14}
\textsc{Slivkins, A.} (2014).
Contextual bandits with similarity information. 
\textit{The Journal of Machine Learning Research} \textbf{15(1)} 2533--2568.

\bibitem{Tho33}
\textsc{Thompson, W.} (1933).
On the likelihood that one unknown probability exceeds another in view of the evidence of two samples.
\textit{Biometrika} \textbf{25} 285--294. 

\bibitem{TG13}
\textsc{Tyagi, H.} and \textsc{G\"artner, B.} (2013). 
Continuum armed bandit problem of few variables in high dimensions. 
\textit{In International Workshop on Approximation and Online Algorithms} 108--119. 

\bibitem{VM05}
\textsc{Vermorel, J.} and \textsc{Mohri, M.} (2005).
Multi-armed bandit algorithms and empirical evaluation.
\textit{Machine Learning: ECML}, Springer, Berlin.
 
\bibitem{Wang2008} \textsc{Wang, Y., Audibert, J.} and \textsc{Munos, R.} (2008). 
Algorithms for infinitely many-armed Bandits. 
\textit{Neural Information Processing Systems}.
	
\end{thebibliography}
\end{document}


\renewcommand{\baselinestretch}{2}

\fontsize{12}{14pt plus.8pt minus .6pt}\selectfont \vspace{0.8pc}
\centerline{\large\bf INFINITE ARMS BANDIT:}
\vspace{2pt} \centerline{\large\bf OPTIMALITY VIA CONFIDENCE BOUNDS }
\vspace{.4cm} \centerline{Hock Peng Chan and Hu Shouri} \vspace{.4cm} \centerline{\it
	National University of Singapore} \vspace{.4cm} \fontsize{9}{11.5pt plus.8pt minus
	.6pt}\selectfont
\fontsize{12}{14pt plus.8pt minus .6pt}\selectfont 
\centerline{\bf Supplementary Material} \vspace{.55cm}

\fontsize{12}{14pt plus.8pt minus .6pt}\selectfont

This document consists of three sections.
In Section A we prove Lemma~2,
in Section B we prove Theorem 1 and in Section C we prove Theorem 2.

\section{Proof of Lemma 2}

Let the infinite arms bandit problem be labeled as Problem A,
and let $R_A$ be the smallest possible regret for this problem.
We prove Lemma 2 by considering two related problems,
Problems B and~C.

\smallskip
{\sc Proof of Lemma 2}. 
Let Problem B be like Problem A except that when we observe the first positive loss from arm $k$,
its mean $\mu_k$ is revealed.
Let $R_B$  be the smallest regret for Problem B.
Since in Problem B we have access to additional arm-mean information,
$R_A \geq R_B$.

In Problem B the best solution involves an initial exploration phase in which we play $K$ arms, each until its first positive loss. 
This is followed by an exploitation phase in which we play the best arm for the remaining $n-M$ trials,
where $M$ is the number of rewards in the exploration phase. 
It is always advantageous to experiment first because no information on arm mean is gained during exploitation.
For continuous rewards $M=K$.
Let $\mu_b(=\mu_{\rm best})=\min_{1 \leq k \leq K} \mu_k$.

In Problem C like in Problem B,
$\mu_k$ is revealed upon the observation of its first positive $X_{kt}$.
The difference is that instead of playing the best arm for $n-M$ additional trials,
we play it for $n$ additional trials,
for a total of $n+M$ trials. 
Let $R_C$ be the smallest regret of Problem C,
the expected value of $\sum_{k=1}^K n_k \mu_k$, 
with $\sum_{k=1}^K n_k=n+M$.
We can extend the optimal solution of Problem B to a (possibly non-optimal) solution of Problem~C by simply playing the best arm with mean $\mu_b$ a further $M$ times.
Hence
\begin{equation} \label{RBC}
[R_A+E(M \mu_b) \geq] R_B+E(M \mu_b) \geq R_C.
\end{equation}
Lemma 2 follows from Lemmas \ref{lem2} and \ref{lem3} below. 
$\wbox$

\begin{customlemma}{3} \label{lem2}
$R_C = n \zeta_n$ for $\zeta_n$ satisfying $v(\zeta_n)= \tfrac{\lambda}{n}$.
\end{customlemma}

\begin{customlemma}{4} \label{lem3}
$E(M \mu_b) = o(n^{\frac{\beta}{\beta+1}})$.
\end{customlemma}

Bonald and Prouti\`{e}re (2013) also referred to Problem B in their lower bounds for Bernoulli rewards.
What is different in our proof of Lemma 2 is a further simplification by considering Problem C,
in which the number of rewards in the exploitation phase is fixed to be $n$.
We show in Lemma \ref{lem2} that under Problem C the optimal regret has a simple expression $n \zeta_n$,
and reduce the proof of Lemma 2 to showing Lemma~\ref{lem3}.

\medskip
{\sc Proof of Lemma} \ref{lem2}. 
Let arm $j$ be the best arm after $k$ arms have been played in the experimentation phase,
that is $\mu_j = \min_{1 \leq i \leq k} \mu_i$.
Let $\phi_*$ be the strategy of trying out a new arm if and only if $n v(\mu_j) > \lambda$,
or equivalently $\mu_j > \zeta_n$.
Since we need on the average $\frac{1}{p(\zeta_n)}$ arms before achieving $\mu_j \leq \zeta_n$,
and the exploration cost of each arm is $\lambda$,
the regret of $\phi_*$ is
\begin{equation} \label{Rstar} 
R_* = \tfrac{\lambda}{p(\zeta_n)}+n E_g(\mu|\mu \leq \zeta_n) = r_n(\zeta_n) = n \zeta_n,
\end{equation}
see (5.1) and Lemma 1 in the main manuscript for the second and third equalities in (\ref{Rstar}).

Hence $R_C \leq n \zeta_n$ and to show Lemma \ref{lem2},
it remains to show that for any strategy $\phi$,
its regret $R_{\phi}$ is not less than $R_*$.
Let $K_*$ be the number of arms played by $\phi_*$ and $K$ the number of arms played by $\phi$.
Let $\mu_* =\min_{1 \leq k \leq K_*} \mu_k$.
Let $G_1 = \{ K < K_* \} (=\{ \min_{1 \leq k \leq K} \mu_k > \zeta_n \})$ and $G_2 = \{ K > K_* \}(=\{ \mu_* \leq \zeta_n, K > K_* \})$.
Since
\begin{eqnarray*}
R_{\phi} & = & \lambda E(K) + n E(\min_{1 \leq k \leq K} \mu_k), \cr
R_* & = & \lambda E(K_*) + n E(\mu_*),
\end{eqnarray*}
we can express
\begin{equation} \label{RR}
R_{\phi}-R_* = \sum_{\ell=1}^2 \Big\{ \lambda E[(K-K_*) {\bf 1}_{G_{\ell}}]+n E \Big[ \Big( \min_{1 \leq k \leq K} \mu_k - \mu_* \Big) {\bf 1}_{G_{\ell}} \Big] \Big\}.
\end{equation}

Under $G_1$,
$\min_{1 \leq k \leq K} \mu_k > \zeta_n$ and therefore by (\ref{Rstar}),
\begin{eqnarray} \label{eq3.1}
& & \lambda E[(K-K_*) {\bf 1}_{G_1}]+ nE \Big[ \Big(\min_{1 \leq k \leq K} \mu_k - \mu_* \Big) {\bf 1}_{G_1} \Big] \\ \nonumber
& = & -\tfrac{\lambda P(G_1)}{p(\zeta_n)}+n \Big\{ E \Big[ \Big( \min_{1 \leq k \leq K} \mu_k \Big) {\bf 1}_{G_1} \Big]-P(G_1)E_g(\mu|\mu \leq \zeta_n) \Big\} \\ \nonumber
& \geq & P(G_1) \{ -\tfrac{\lambda}{p(\zeta_n)}+n[\zeta_n-E_g(\mu|\mu \leq \zeta_n)] \} = 0. 
\end{eqnarray}
The identity $E[(K_*-K) {\bf 1}_{G_1}] = \tfrac{P(G_1)}{p(\zeta_n)}$ is due to $\min_{1 \leq k \leq K} \mu_k > \zeta_n$ when there are $K$ arms,
and so an additional $\frac{1}{p(\zeta_n)}$ arms on average is required under strategy $\phi_*$,
to get an arm with $\mu_k$ not more than $\zeta_n$.
The identity
$$E(\mu_* {\bf 1}_{G_1}) = P(G_1) E(\mu_*) = P(G_1) E_g(\mu|\mu \leq \zeta_n)
$$
is due to the independence between ${\bf 1}_{G_1}$ and $\mu_*$.

In view that $(K-K_*) {\bf 1}_{G_2} =\sum_{j=0}^{\infty} {\bf 1}_{\{ K > K_*+j \}}$ and
\begin{eqnarray*}
& & \Big( \min_{1 \leq k \leq K} \mu_k - \mu_* \Big) {\bf 1}_{G_2} \cr
& = & \sum_{j=0}^{\infty} \Big( \min_{1 \leq k \leq K_*+j+1} \mu_k - \min_{1 \leq k \leq K_*+j} \mu_k \Big) {\bf 1}_{\{ K > K_*+j \}},
\end{eqnarray*}
it follows that
\begin{eqnarray} \label{eq3.2}
& & \lambda E[(K-K_*) {\bf 1}_{G_2}]+n E \Big[ \Big( \min_{1 \leq k \leq K} \mu_k-\mu_* \Big) {\bf 1}_{G_2} \Big] \\ \nonumber
& = & \sum_{j=0}^{\infty} E \Big\{ \Big[ \lambda + n \Big( \min_{1 \leq k \leq K_*+j+1} \mu_k - \min_{1 \leq k \leq K_*+j} \mu_k \Big) \Big] {\bf 1}_{\{ K > K_*+j \}} \Big\} \\ \nonumber
& = & \sum_{j=0}^{\infty} E \Big\{ \Big[ \lambda-n v \Big(\min_{1 \leq k \leq K_*+j} \mu_k \Big) \Big] {\bf 1}_{ \{ K>K_*+j \}} \Big\} \geq 0.
\end{eqnarray}
The second equality in (\ref{eq3.2}) follows from
$$E \Big( \min_{1 \leq k \leq K_*+j} \mu_k - \min_{1 \leq k \leq K^*+j+1} \mu_k \Big| \min_{1 \leq k \leq K^*+j} \mu_k=x, K > K^*+j \Big) = v(x).
$$
The inequality in (\ref{eq3.2}) follows from 
$$v \Big( \min_{1 \leq k \leq K_*+j} \mu_k \Big) \leq v(\mu_*) \leq v(\zeta_n) = \tfrac{\lambda}{n},
$$
as $v$ is monotone increasing. 
Lemma \ref{lem2} follows from (\ref{Rstar})--(\ref{eq3.2}).
$\wbox$

\medskip
{\sc Proof of Lemma} \ref{lem3}. 
Let $\wht K = \lfloor n \zeta_n (\log n)^{\beta+2} \rfloor$ for $\zeta_n$ satisfying $n v(\zeta_n)=\lambda$.
Express $E(M \mu_b) = \sum_{i=1}^5 E(M \mu_b {\bf 1}_{D_i})$, 
where 
\begin{eqnarray*}
	D_1 & = & \{ \mu_b \leq \tfrac{\zeta_n}{\log n} \}, \cr
	D_2 & = & \{ \mu_b > \tfrac{\zeta_n}{\log n}, K > \wht K \}, \cr
	D_3 & = & \{ \tfrac{\zeta_n}{\log n} < \mu_b \leq \zeta_n (\log n)^{\beta+3}, K \leq \wht K \}, \cr
	D_4 & = & \{ \mu_b > \zeta_n (\log n)^{\beta+3}, K \leq \wht K, M > \tfrac{n}{2} \}, \cr
	D_5 & = & \{ \mu_b > \zeta_n (\log n)^{\beta+3}, K \leq \wht K, M \leq \tfrac{n}{2} \}.
\end{eqnarray*}

It suffices to show that for all $i$,
\begin{equation} \label{ED}
E(M \mu_b {\bf 1}_{D_i}) = o(n^{\frac{\beta}{\beta+1}}).
\end{equation}
Since $\zeta_n \sim Cn^{-\frac{1}{\beta+1}}$ [see (3.3) of the main manuscript],
$\frac{M \zeta_n}{\log n} \leq \frac{n \zeta_n}{\log n} = o(n^{\frac{\beta}{\beta+1}})$ and (\ref{ED}) holds for $i=1$.

Let $\wht \mu_b = \min_{k \leq \hat K} \mu_k$.
Since $M \leq n$, 
$\mu_b \leq \mu_1$ and $E(\mu_1)\leq \lambda$, 
\begin{eqnarray} \label{EM2}
E(M \mu_b {\bf 1}_{D_2}) & \leq & n E(\mu_1 {\bf 1}_{D_2}) \\ \nonumber
& = & n E(\mu_1|\mu_1 > \tfrac{\zeta_n}{\log n}) P(D_2) \\ \nonumber
& \leq & [\lambda+o(1)] n P(\wht \mu_b > \tfrac{\zeta_n}{\log n}).
\end{eqnarray}
By condition (A1),
$p(\zeta) \sim \frac{\alpha}{\beta} \zeta^{\beta}$ as $\zeta \rightarrow 0$,
hence substituting
$$P(\wht \mu_b > \tfrac{\zeta_n}{\log n}) = [1-p(\tfrac{\zeta_n}{\log n})]^{\hat K} = \exp \{ -[1+o(1)] \wht K \tfrac{\alpha}{\beta} (\tfrac{\zeta_n}{\log n})^{\beta}] = O(n^{-1})
$$
into (\ref{EM2}) shows (\ref{ED}) for $i=2$.

Let $M_j$ be the number of plays of $\Pi_j$ to the first positive $X_{jt}$ (hence $M = \sum_{j=1}^K M_j$).
It follows from condition (A2) that $E_{\mu} M_1 = \frac{1}{P_{\mu}(X_1>0)} \leq \frac{1}{a_1 \min(\mu,1)}$,
hence by $\mu_b \leq \zeta_n (\log n)^{\beta+3}$ under $D_3$,
\begin{eqnarray} \label{D3}
E(M \mu_b {\bf 1}_{D_3}) & \leq & E(M_1 {\bf 1}_{\{ \mu_1 > \frac{\zeta_n}{\log n} \}}) \wht K \zeta_n (\log n)^{\beta+3} \\ \nonumber 
& \leq & \Big( \int_{\frac{\zeta_n}{\log n}}^{\infty} \tfrac{g(\mu)}{a_1 \min(\mu,1)} d \mu \Big) n \zeta_n^2 (\log n)^{2 \beta+5}.
\end{eqnarray}
Substituting 
$$\int_{\frac{\zeta_n}{\log n}}^\infty \tfrac{g(\mu)}{\mu} d \mu = \left\{ \begin{array}{ll} O(1) & \mbox{ if } \beta > 1, \cr
O(\log n) & \mbox{ if } \beta=1, \cr
O((\tfrac{\zeta_n}{\log n})^{\beta-1}) & \mbox{ if } \beta<1, \end{array} \right.
$$
into (\ref{D3}) shows (\ref{ED}) for $i=3$.

If $\mu_j > \zeta_n (\log n)^{\beta+3}$,
then by condition (A2),
$M_j$ is bounded above by a geometric random variable with mean $\nu^{-1}$,
where $\nu = a_1 \zeta_n (\log n)^{\beta+3}$.
Hence for $0 < \theta < \log(\frac{1}{1-\nu})$,
$$E(e^{\theta M_j} {\bf 1}_{\{ \mu_j > \zeta_n (\log n)^{\beta+3} \}}) 
\leq \sum_{h=1}^{\infty} e^{\theta h} \nu (1-\nu)^{h-1} = \tfrac{\nu e^{\theta}}{1-e^{\theta}(1-\nu)}, 
$$
implying that
\begin{equation} \label{EtM}
[e^{\frac{\theta n}{2}} P(D_4) \leq] E (e^{\theta M} {\bf 1}_{D_4}) \leq (\tfrac{\nu e^{\theta}}{1-e^{\theta}(1-\nu)})^{\hat K}.
\end{equation}
Consider $\theta$ such that $e^{\theta}=1+\frac{\nu}{2}$ and check that $e^{\theta}(1-\nu) \leq 1-\frac{\nu}{2}$ [$\Rightarrow \theta < \log(\frac{1}{1-\nu})]$.
It follows from (\ref{EtM}) that
\begin{eqnarray*}
	P(D_4) & \leq & e^{-\frac{\theta n}{2}} (\tfrac{\nu e^{\theta}}{\nu/2})^{\hat K} = 2^{\hat K} e^{\theta(\hat K-\frac{n}{2})} \cr
	& = & \exp[\wht K \log 2+[1+o(1)] \tfrac{\nu}{2}(\wht K-\tfrac{n}{2})]  \cr
    & = &  \exp \{-[1+o(1)] \tfrac{n \nu}{4} \} = O(n^{-1}).	
\end{eqnarray*}
Since $M \leq n$, $\mu_b \leq \mu_1$ and $E(\mu_1) \leq \lambda$, 
$$E(M \mu_b {\bf 1}_{D_4}) \leq n E[\mu_1|\mu_1 > \zeta_n (\log n)^{\beta+3}] P(D_4) \leq n[\lambda+o(1)] P(D_4),
$$
and (\ref{ED}) holds for $i=4$.

Under $D_5$ for $n$ large,
since $v(\zeta) \sim \tfrac{\alpha}{\beta(\beta+1)} \zeta^{\beta+1}$ as $\zeta \rightarrow 0$ and $\zeta_n \sim Cn^{-\frac{1}{\beta+1}}$,
$$(n-M) v(\mu_b) [> \tfrac{n}{2} v(\zeta_n (\log n)^{\beta+3})] > \lambda.
$$
If we explore one more arm,
then the additional exploration cost is not more than $\lambda$ and reduction in exploitation cost is at least $(n-K) v(\mu_b)$.
Hence $D_5$ is an event of zero probability,
in view that we are looking at the optimal solution of Problem B.
Therefore (\ref{ED}) holds for $i=5$.
$\wbox$

\section{Proof of Theorem 1}

We preface the proof of Theorem 1 with Lemmas \ref{lem4}--\ref{lem7}.
The lemmas are proved in Section B.1 and B.2.
Consider $X_1, X_2, \ldots$ i.i.d. $F_{\mu}$.
Let $S_t = \sum_{u=1}^t X_u$,
$\bar X_t = \frac{S_t}{t}$ and $\wht \sigma_t^2 = t^{-1} \sum_{u=1}^t (X_u-\bar X_t)^2$.
Let
\begin{eqnarray} \label{T1}
T_b  & = & \inf \{ t: S_t > b_n t \zeta_n \}, \\ \label{T2}
T_c & = & \inf \{ t: S_t > t \zeta_n + c_n \wht \sigma_t \sqrt{t} \},  
\end{eqnarray}
with $b_n \rightarrow \infty$ and $c_n \rightarrow \infty$ such that $b_n+c_n=o(n^{\delta})$ for all $\delta>0$,
and $\zeta_n \sim Cn^{-\frac{1}{\beta+1}}$ for $C=(\frac{\lambda \beta(\beta+1)}{\alpha})^{\frac{1}{\beta+1}}$. 
Let 
\begin{equation}
\label{dn}
d_n = n^{-\omega} \text{ for some }  0 < \omega < \tfrac{1}{\beta+1}.
\end{equation}

\begin{customlemma}{5} \label{lem4}
As $n \rightarrow \infty$,
\begin{eqnarray} \label{lem4.1}
\sup_{\mu \geq d_n} [\min(\mu,1) E_{\mu} T_b] & = & O(1), \\ \label{lem4.2}
E_g(T_b \mu {\bf 1}_{\{ \mu \geq d_n \}}) & \leq & \lambda+o(1). 
\end{eqnarray}
\end{customlemma}

\begin{customlemma}{6} \label{lem5}
Let $\epsilon >0$.
As $n \rightarrow \infty$,
\begin{eqnarray} \label{lem5.1}
\sup_{(1+\epsilon)\zeta_n \leq \mu \leq d_n} [\mu E_{\mu}(T_c \wedge n)] & = & O(c_n^3+\log n), \\ \label{lem5.2}
E_g[(T_c \wedge n) \mu {\bf 1}_{\{ (1+\epsilon)\zeta_n \leq \mu \leq d_n \}}] & \rightarrow & 0.
\end{eqnarray}
\end{customlemma}

\begin{customlemma}{7} \label{lem6}
Let $0 < \epsilon < 1$.
As $n \rightarrow \infty$, 
$$\sup_{\mu \leq (1-\epsilon)\zeta_n} P_{\mu}(T_b < \infty) \rightarrow 0.
$$
\end{customlemma}

\begin{customlemma}{8} \label{lem7}
Let $0 < \epsilon < 1$.
As $n \rightarrow \infty$,
$$\sup_{\mu \leq (1-\epsilon)\zeta_n} P_{\mu}(T_c < \infty) \rightarrow 0.
$$
\end{customlemma}

The number of times an arm is played has distribution bounded above by $T:= T_b \wedge T_c$.
Lemmas \ref{lem6} and \ref{lem7} say that an arm with $\mu_k$ less than $(1-\epsilon) \zeta_n$ is unlikely to be rejected,
whereas (\ref{lem4.2}) and (\ref{lem5.2}) say that the regret due to sampling from an arm with $\mu_k$ more than $(1+\epsilon) \zeta_n$ is asymptotically bounded by $\lambda$.
The remaining (\ref{lem4.1}) and (\ref{lem5.1}) are technical relations used in the proof of Theorem 1.

\smallskip
{\sc Proof of Theorem} 1.
The number of times arm $k$ is played is $n_k$,
and it is distributed as $T_b\wedge T_c \wedge (n-\sum_{\ell=1}^{k-1} n_{\ell})$. Let $0 <\epsilon <1$. We can express
\begin{equation} \label{Rnz}
R_n - n \zeta_n = z_1+z_2+z_3 = z_1+z_2-|z_3|, 
\end{equation}
where $z_i = E[\sum_{k: \mu_k \in D_i} n_k(\mu_k-\zeta_n)]$ for
$$D_1 = [(1+\epsilon) \zeta_n,\infty), \quad D_2 = ((1-\epsilon) \zeta_n,(1+\epsilon) \zeta_n), \quad D_3 = (0,(1-\epsilon) \zeta_n].
$$
It is easy to see that $z_2 \leq \epsilon n\zeta_n$. 
We shall show that
\begin{eqnarray} \label{z1}
z_1 & \leq & \tfrac{\lambda+o(1)}{(1-\epsilon)^{\beta} p(\zeta_n)}, \\ \label{z3m}
|z_3| & \geq & [(\tfrac{1-\epsilon}{1+\epsilon})^{\beta}+o(1)][n \epsilon \zeta_n+\tfrac{(1-\epsilon)\lambda}{p(\zeta_n)}].
\end{eqnarray}
We conclude Theorem 1 from (\ref{Rnz})--(\ref{z3m}) with $\epsilon \to 0$.
$\wbox$

\smallskip
{\sc Proof of} (\ref{z1}). 
Since $T=T_b \wedge T_c$,
by Lemmas~\ref{lem6} and \ref{lem7},
\begin{eqnarray}\label{qn}
q_n & := & \sup_{\mu \leq (1-\epsilon) \zeta_n} P_{\mu}(T < \infty)\\ \nonumber
 &\leq & \sup_{\mu \leq (1-\epsilon) \zeta_n}[P_{\mu}(T_b < \infty)+P_{\mu}(T_c < \infty)] \rightarrow 0.
\end{eqnarray}
That is an arm with $\mu_k$ less than $(1-\epsilon) \zeta_n$ is rejected with negligible probability for $n$ large. 
Since the total number of played arms $K$ is bounded above by a geometric random variable with mean $\frac{1}{P_g(T=\infty)}$, 
by (\ref{qn}) and $p(\zeta) \sim \frac{\alpha}{\beta} \zeta^{\beta}$ as $\zeta \rightarrow 0$,
\begin{equation}
\label{EK}
EK \leq \tfrac{1}{P_g(T=\infty)} \leq \tfrac{1}{(1-q_n)p((1-\epsilon) \zeta_n)} \sim \tfrac{1}{(1-\epsilon)^\beta p(\zeta_n)}.
\end{equation}
By (\ref{lem4.2}) and (\ref{lem5.2}),
\begin{eqnarray*}
& & E_g(n_1 \mu_1 {\bf 1}_{\{ \mu_1 \geq (1+\epsilon) \zeta_n \}}) \cr
& = & E_g(n_1 \mu_1 {\bf 1}_{\{(1+\epsilon) \zeta_n \leq \mu_1 \leq d_n\}})+E_g(n_1 \mu_1 {\bf 1}_{\{ \mu_1 \geq d_n \}}) \cr
& \leq & E_g[(T_c \wedge n) \mu_1 {\bf 1}_{\{(1+\epsilon) \zeta_n \leq \mu_1 \leq d_n\}}]+E_g(T_b \mu_1 {\bf 1}_{\{ \mu_1 \geq d_n \}}) \cr
& \leq & \lambda+o(1),
\end{eqnarray*}
and (\ref{z1}) follows from (\ref{EK}) and $z_1 \leq E_g(n_1 \mu_1 {\bf 1}_{\{ \mu_1 \geq (1+\epsilon) \zeta_n \}}) EK$. 
$\wbox$

\smallskip
{\sc Proof of} (\ref{z3m}).
Let $\ell$ be the first arm with mean not more than $(1-\epsilon) \zeta_n$.
We have
\begin{eqnarray} \label{|z3|}
|z_3| & = & E \Big[ \sum_{k: \mu_k \in D_3} n_k(\zeta_n-\mu_k) \Big] \\ \nonumber
& \geq & (E n_{\ell}) \{ \zeta_n - E_g[\mu|\mu \leq (1-\epsilon) \zeta_n] \}.
\end{eqnarray}
Since $v(\zeta_n) \sim \frac{\lambda}{n}$ and $p(\zeta) \sim \frac{\alpha}{\beta} \zeta^{\beta}$,
$v(\zeta) \sim \frac{\alpha}{\beta(\beta+1)} \zeta^{\beta+1}$ as $\zeta \rightarrow 0$,
\begin{eqnarray*} 
& & \zeta_n-E_g[\mu|\mu \leq (1-\epsilon) \zeta_n] \cr
& = & \zeta_n-\{ (1-\epsilon) \zeta_n-E_g[(1-\epsilon) \zeta_n-\mu| \mu \leq (1-\epsilon) \zeta_n] \} \cr
& = & \zeta_n-[(1-\epsilon)\zeta_n-\tfrac{v((1-\epsilon) \zeta_n)}{p((1-\epsilon) \zeta_n)}] \cr
& \sim & \epsilon \zeta_n + \tfrac{(1-\epsilon) v(\zeta_n)}{p(\zeta_n)} \sim \epsilon \zeta_n + \tfrac{(1-\epsilon) \lambda}{n p(\zeta_n)},
\end{eqnarray*}
and (\ref{z3m}) thus follows from (\ref{|z3|}) and 
\begin{equation} \label{Enl}
E n_{\ell} \geq [(\tfrac{1-\epsilon}{1+\epsilon})^{\beta}+o(1)] n.
\end{equation}

Let $j$ be the first arm with mean not more than $(1+\epsilon) \zeta_n$ and $M = \sum_{i=1}^{j-1} n_i$.
We have
$$E n_{\ell} \geq (1-q_n) E(n-M) P(\ell=j).
$$
Since $q_n \rightarrow 0$ and $P(\ell=j) \rightarrow (\frac{1-\epsilon}{1+\epsilon})^{\beta}$,
to show (\ref{Enl}) it suffices to show that $EM=o(n)$.

Indeed by (\ref{lem4.1}),
(\ref{lem5.1}) and $E_{\mu} n_1 \leq E_{\mu}(T \wedge n)$, 
\begin{eqnarray*} 
& & \sup_{\mu \geq (1+\epsilon) \zeta_n} [\min(\mu,1) E_{\mu} n_1] \cr
& \leq & \max \Big[ 	\sup_{(1+\epsilon) \zeta_n \leq \mu \leq d_n} \mu E_\mu(T_c \wedge n), \sup_{\mu \geq d_n} \min(\mu,1) E_{\mu}T_b \Big]	= O(c_n^3+\log n).
\end{eqnarray*}
Hence in view that $\frac{1}{p((1+\epsilon)\zeta_n)} = O(n^{\frac{\beta}{\beta+1}})$ and $P_g(\mu_1 > (1+\epsilon) \zeta_n) \rightarrow 1$ as $n \rightarrow \infty$,
\begin{eqnarray*} 
EM & \leq & \tfrac{1}{p((1+\epsilon) \zeta_n)} E_g(n_1|\mu_1 > (1+\epsilon) \zeta_n) \cr
& = & O( n^{\frac{\beta}{\beta+1}}) E_g[\tfrac{c_n^3+\log n}{\min(\mu_1,1)} \big| \mu_1 > (1+\epsilon) \zeta_n ] \cr
& = & O( n^{\frac{\beta}{\beta+1}} (c_n^3+\log n)) \int_{(1+\epsilon) \zeta_n}^{\infty} \tfrac{g(\mu)}{\min(\mu,1)} d \mu \cr
& = & O( n^{\frac{\beta}{\beta+1}} (c_n^3+\log n)) \max(n^{\frac{1-\beta}{\beta+1}},\log n) = o(n). 
\end{eqnarray*}
The first relation in the line above follows from
$$\int_{(1+\epsilon)\zeta_n}^\infty \tfrac{g(\mu)}{\min(\mu,1)} d \mu = \left\{ \begin{array}{ll} O(1) & \mbox{ if } \beta > 1, \cr
O(\log n) & \mbox{ if } \beta=1, \cr
O(n^{\frac{1-\beta}{\beta+1}}) & \mbox{ if } \beta<1. \quad \mbox{$\wbox$}  \end{array} \right. 
$$

\subsection{Proofs of Lemmas \ref{lem4}--\ref{lem7} for discrete rewards}

In the case of discrete rewards,
one difficulty is that for $\mu_k$ small,
there are potentially multiple plays on arm $k$ before a positive $X_{kt}$ is observed.
Condition (A2) is helpful in ensuring that the mean of this positive $X_{kt}$ is not too large.

Recall that for integer-valued rewards we assume in condition (B1) that for $0 < \delta \leq 1$,
there exists $\theta_{\delta}>0$ such that for $\mu>0$ and $0 \leq \theta \leq \theta_{\delta}$,
\begin{eqnarray} \label{Mbound}
M_{\mu}(\theta) & \leq & e^{(1+\delta) \theta \mu}, \\ \label{Mmu2}
M_{\mu}(-\theta) & \leq & e^{-(1-\delta) \theta \mu}.
\end{eqnarray}
In addition,
\begin{eqnarray} \label{a2}
P_{\mu}(X>0) & \leq & a_2 \mu \mbox{ for some } a_2>0, \\ \label{E4}
E_{\mu} X^4 & = & O(\mu) \mbox{ as } \mu \rightarrow 0.
\end{eqnarray}

\smallskip
{\sc Proof of Lemma} \ref{lem4}.
Recall that
$$T_b = \inf \{ t: S_t > b_n t \zeta_n \},  
$$
and that $d_n = n^{-\omega}$ for some $0 < \omega < \frac{1}{\beta+1}$.
We shall show that 
\begin{eqnarray} \label{ET}
\sup_{\mu \geq d_n} [\min(\mu,1) E_{\mu} T_b] & = & O(1), \\ \label{limlam}
E_g(T_b \mu {\bf 1}_{\{ \mu \geq d_n \}}) & \leq & \lambda+o(1).
\end{eqnarray}

Let $\theta = 2\omega \log n$.
Since $X_u$ is integer-valued,
it follows from Markov's inequality that 
\begin{equation} \label{Stb}
P_{\mu}(S_t \leq b_n t \zeta_n) \leq [e^{\theta b_n \zeta_n} M_{\mu}(-\theta)]^t \leq \{ e^{\theta b_n \zeta_n} [P_{\mu}(X=0)+e^{-\theta}] \}^t.
\end{equation}

By $P_{\mu}(X>0) \geq a_1 d_n$ for $\mu \geq d_n$ [see (A2)],
$\theta b_n \zeta_n = o(d_n)$ [because $\theta$ and $b_n$ are both sub-polynomial in $n$ and $\zeta_n = O(n^{-\frac{1}{\beta+1}})]$ and (\ref{Stb}),
uniformly over $\mu \geq d_n$,
\begin{eqnarray} \label{ET1}
E_{\mu} T_b & = & 1 + \sum_{t=1}^{\infty} P_{\mu}(T_b>t) \\ \nonumber
& \leq & 1+ \sum_{t=1}^{\infty} P_{\mu}(S_t \leq b_n t \zeta_n) \\ \nonumber
& \leq & \{ 1-e^{\theta b_n \zeta_n}[P_{\mu}(X=0)+e^{-\theta}] \}^{-1} \\ \nonumber
& = & \{ 1-[1+o(d_n)][P_{\mu}(X=0)+d_n^2] \}^{-1} \\ \nonumber
& = & [ P_{\mu}(X>0)+o(d_n)]^{-1} \sim [P_{\mu}(X>0)]^{-1}. 
\end{eqnarray}
The term inside $\{ \cdot \}$ in (\ref{Stb}) is not more than $[1+o(d_n)](1-a_1 d_n+d_n^2) < 1$ for $n$ large and this gives us the second inequality in (\ref{ET1}).
We conclude (\ref{ET}) from (\ref{ET1}) and (A2). 
By (\ref{ET1}), 
\begin{eqnarray*}
	E_g[T_b \mu {\bf 1}_{\{ \mu \geq d_n \}}] & = & \int_{d_n}^{\infty} E_{\mu}(T_b)\mu  g(\mu) d \mu \cr
	& \leq  & [1+o(1)] \int_{d_n}^{\infty} \tfrac{E_{\mu}(X)}{P_{\mu}(X>0)} g(\mu) d \mu \cr
	& = & [1+o(1)] \int_{d_n}^{\infty} E_{\mu}(X|X>0) g(\mu) d \mu \rightarrow \lambda, 
\end{eqnarray*}
hence (\ref{limlam}) holds. 
$\wbox$

\smallskip
{\sc Proof of Lemma} \ref{lem5}.
Recall that $T_c = \inf \{ t: S_t > t \zeta_n + c_n \wht \sigma_t \sqrt{t} \}$ and let $\epsilon > 0$. 
We want to show that 
\begin{eqnarray} \label{lem5a}
\sup_{(1+\epsilon)\zeta_n \leq\mu \leq d_n} \mu E_{\mu} (T_c \wedge n) & =&  O(c_n^3+ \log n), \\ \label{lem5b} 
E_g[(T_c \wedge n) \mu {\bf 1}_{\{(1+\epsilon)\zeta_n \leq \mu \leq d_n \}}] & \rightarrow & 0.
\end{eqnarray}

We first show that there exists $\kappa>0$ such that as $n \rightarrow \infty$,
\begin{equation} \label{show}
\sup_{\mu \leq d_n} \Big[ \mu \sum_{t=1}^n P_{\mu}(\wht \sigma_t^2 \geq \kappa \mu) \Big] =O(\log n).
\end{equation}
Since $X$ is non-negative integer-valued, $X^2 \leq X^4$. Indeed by (\ref{E4}),
there exists $\kappa>0$ such that $\rho_{\mu}:=E_{\mu} X^2 \leq \frac{\kappa \mu}{2}$ for $\mu \leq d_n$ and $n$ large,
therefore by (\ref{E4}) again and Chebyshev's inequality,
\begin{eqnarray*}
	P_{\mu}(\wht \sigma_t^2 \geq \kappa \mu) & \leq & P_{\mu} \Big( \sum_{u=1}^t X_u^2 \geq t \kappa \mu \Big) \cr
	& \leq & P_{\mu} \Big( \sum_{u=1}^t (X_u^2-\rho_{\mu}) \geq \tfrac{t \kappa \mu}{2} \Big) \cr
	& \leq & \tfrac{t {\rm Var}_{\mu}(X^2)}{(t \kappa \mu/2)^2} = O((t \mu)^{-1}), 
\end{eqnarray*}
and (\ref{show}) holds. 

By (\ref{show}),
uniformly over $(1+\epsilon)\zeta_n \leq \mu \leq d_n$,
\begin{eqnarray} \label{En2}
E_{\mu} (T_c \wedge n) & = & 1 + \sum_{t=1}^{n-1} P_{\mu}(T_c > t) \\ \nonumber
& \leq & 1+ \sum_{t=1}^{n-1} P_{\mu}(S_t \leq t \zeta_n + c_n \wht \sigma_t\sqrt{t})  \\ \nonumber
& \leq & 1+ \sum_{t=1}^{n-1} P_{\mu}(S_t \leq t \zeta_n + c_n \sqrt{\kappa\mu t}) + O(\tfrac{\log n}{\mu}).
\end{eqnarray}

Let $0 < \delta < \frac{1}{2}$ to be further specified.
Uniformly over $t \geq c_n^3 \mu^{-1}$,
$\mu t/(c_n \sqrt{\kappa\mu t}) \rightarrow \infty$ and therefore by (\ref{Mmu2}), $\mu \geq (1+\epsilon)\zeta_n $ and Markov's inequality, 
for $n$ large,
\begin{eqnarray} \label{case2}
P_{\mu}(S_t \leq t \zeta_n+c_n \sqrt{\kappa\mu t}) & \leq & P_{\mu}(S_t \leq t(\zeta_n+\delta \mu)) \\ \nonumber
& \leq & e^{\theta_{\delta} t (\zeta_n+\delta \mu)} M_{\mu}^t (-\theta_{\delta}) \\ \nonumber
& \leq & e^{t \theta_{\delta}[\zeta_n-(1-2 \delta) \mu]} \leq e^{-\eta t \theta_{\delta} \mu},
\end{eqnarray}
where $\eta =1-2 \delta-\frac{1}{1+\epsilon} >0$ (for $\delta$ chosen small).
Since $1-e^{-\eta \theta_{\delta} \mu} \sim \eta \theta_{\delta} \mu$ as $\mu \rightarrow 0$,
\begin{equation} \label{cn3}
\sum_{t=1}^{n-1} e^{-\eta t \theta_{\delta} \mu} \leq c_n^3 \mu^{-1} + \sum_{t \geq c_n^3 \mu^{-1}} e^{-\eta t \theta_{\delta} \mu} = O(c_n^3 \mu^{-1}),
\end{equation}
and substituting (\ref{case2}) into (\ref{En2}) gives us (\ref{lem5a}). 
By (\ref{lem5a}),
\begin{eqnarray} \nonumber
E_g[(T_c \wedge n) \mu {\bf 1}_{\{(1+\epsilon)\zeta_n \leq  \mu \leq d_n \}}] & = & P_g((1+\epsilon)\zeta_n \leq  \mu \leq d_n) O(c_n^3+\log n) \\ \nonumber
& = & O(d_n^{\beta}(c_n^3+\log n)),
\end{eqnarray}
and (\ref{lem5b}) holds since $c_n$ is sub-polynomial in $n$. $\wbox$

\smallskip
{\sc Proof of Lemma} \ref{lem6}.
We want to show that 
\begin{equation} \label{Pgn1}
P_{\mu}(S_t > tb_n \zeta_n \mbox{ for some } t \geq 1) \rightarrow 0
\end{equation}
uniformly over $\mu \leq (1-\epsilon)\zeta_n$.

By (\ref{a2}) and Bonferroni's inequality,
\begin{eqnarray} \label{lem6.1}
& & P_{\mu}(S_t > tb_n \zeta_n \mbox{ for some } t \leq \tfrac{1}{\sqrt{b_n} \zeta_n}) \\ \nonumber
& \leq & P_{\mu}(X_t > 0 \mbox{ for some } t \leq \tfrac{1}{\sqrt{b_n} \zeta_n}) \leq \tfrac{a_2 \mu}{\sqrt{b_n} \zeta_n} \rightarrow 0.
\end{eqnarray}
By (\ref{Mbound}) and Markov's inequality,
for $n$ large,
\begin{eqnarray} \label{lem6.2}
& & P_{\mu}(S_t > tb_n \zeta_n \mbox{ for some } t > \tfrac{1}{\sqrt{b_n} \zeta_n}) \\ \nonumber
& \leq & \sup_{t > \frac{1}{\sqrt{b_n} \zeta_n}} [e^{-\theta_1 b_n \zeta_n} M_{\mu}(\theta_1)]^t \leq e^{-\theta_1(b_n \zeta_n-2 \mu)/(\zeta_n \sqrt{b_n})} \rightarrow 0.
\end{eqnarray}

To see the first inequality of (\ref{lem6.2}),
let $f_\mu$ be the density of $X_1$ with respect to some $\sigma$-finite measure, and let $E_\mu^{\theta_1} (P_\mu^{\theta_1})$ denote expectation (probability) with respect to density
\begin{equation*}
f_\mu^{\theta_1}(x) := [M_\mu(\theta_1)]^{-1}e^{\theta_1 x} f_\mu(x).
\end{equation*}
Let $T = \inf\{ t > \tfrac{1}{\sqrt{b_n} \zeta_n} : S_t > tb_n\zeta_n\}$.
It follows from Markov's inequality that
\begin{eqnarray} \label{com}
P_\mu(T = t) & = & M_\mu^t(\theta_1) E_\mu^{\theta_1}(e^{-\theta_1 S_t}\mathbf{1}_{\lbrace T = t \rbrace}) \\ \nonumber
& \leq & [e^{-\theta_1 b_n\zeta_n} M_\mu(\theta_1)]^t P_\mu^{\theta_1}(T=t),
\end{eqnarray}
and the first inequality of (\ref{lem6.2}) follows from summing (\ref{com}) over $t > \tfrac{1}{\sqrt{b_n} \zeta_n}$.  
$\wbox$

\smallskip
{\sc Proof of Lemma} \ref{lem7}.
We want to show that 
\begin{equation} \label{lem7.1}
P_{\mu}(S_t > t \zeta_n + c_n \wht \sigma_t \sqrt{t} \mbox{ for some } t \geq 1) \rightarrow 0
\end{equation}
uniformly over $\mu \leq (1-\epsilon)\zeta_n$.

By (\ref{a2}) and Bonferroni's inequality,
\begin{eqnarray} \label{Bonf}
& & P_{\mu}(S_t > t \zeta_n + c_n \wht \sigma_t \sqrt{t} \mbox{ for some } t \leq \tfrac{1}{c_n \mu}) \\ \nonumber
& \leq & P_{\mu}(X_t>0 \mbox{ for some } t \leq \tfrac{1}{c_n \mu}) \leq \tfrac{a_2}{c_n} \rightarrow 0.
\end{eqnarray}
Moreover
\begin{equation} \label{lem7.2}
P_{\mu}(S_t > t \zeta_n+c_n \wht \sigma_t \sqrt{t} \mbox{ for some } t > \tfrac{1}{c_n \mu}) \leq {\rm (I)} + {\rm (II)},
\end{equation}
\begin{eqnarray} \nonumber
\mbox{where (I)} & = & P_{\mu}(S_t > t \zeta_n+c_n (\mu t/2)^{\frac{1}{2}} \mbox{ for some } t > \tfrac{1}{c_n \mu}), \\ \nonumber
{\rm (II)} & = & P_{\mu}(\wht \sigma_t^2 \leq \tfrac{\mu}{2} \mbox{ and } S_t \geq t \zeta_n \mbox{ for some } t > \tfrac{1}{c_n \mu}).
\end{eqnarray} 
By (\ref{Bonf}) and (\ref{lem7.2}),
to show (\ref{lem7.1}),
it suffices to show that (I)$\rightarrow 0$ and (II)$\rightarrow 0$.

Let $0 < \delta \leq 1$ be such that $1+\delta < (1-\epsilon)^{-1}$.
Hence $\mu \leq (1-\epsilon)\zeta_n$ implies $\zeta_n \geq (1+\delta) \mu$.
It follows from (\ref{Mbound}) and Markov's inequality [see (\ref{lem6.2}) and (\ref{com})] that 
\begin{eqnarray*}
	{\rm (I)} & \leq & \sup_{t>\frac{1}{c_n \mu}} [e^{-\theta_{\delta}[t \zeta_n+c_n(\mu t/2)^{\frac{1}{2}}]} M_{\mu}^t(\theta_{\delta})] \\
	& \leq & \exp \{ -\theta_{\delta}[\zeta_n-(1+\delta) \mu]/(c_n \mu)-\theta_{\delta} (c_n/2)^{\frac{1}{2}} \}\\
	& \leq & \exp \{-\theta_{\delta} (c_n/2)^{\frac{1}{2}}\} \rightarrow 0.
\end{eqnarray*}
Since $X_u^2 \geq X_u$,
the inequality $S_t \geq t \zeta_n (\geq t \mu)$ implies $\sum_{u=1}^t X_u^2 \geq t \mu$,
and this,
together with $\wht \sigma_t^2 \leq \frac{\mu}{2}$ implies that $\bar X_t^2 \geq \frac{\mu}{2}$.
Hence by (\ref{Mbound}) and Markov's inequality argument,
for $n$ large,
\begin{eqnarray*}
	\mbox{(II)} & \leq & P_{\mu}(\bar X_t \geq \sqrt{\tfrac{\mu}{2}} \mbox{ for some } t > \tfrac{1}{c_n \mu}) \cr
	& \leq & \sup_{t > \frac{1}{c_n \mu}} [e^{-\theta_1 \sqrt{\mu/2}} M_{\mu}(\theta_1)]^t \cr
	& \leq & \exp \{-\theta_1[\sqrt{\tfrac{\mu}{2}}-2 \mu]/(c_n \mu)\} \cr
	&  \leq & \exp \big\{-\theta_1 \big[ \tfrac{1}{c_n \sqrt{2(1-\epsilon)\zeta_n}}-\tfrac{2}{c_n} \big] \big\}  \rightarrow 0 . \quad \mbox{$\wbox$}
\end{eqnarray*}

\subsection{Proofs of Lemmas \ref{lem4}--\ref{lem7} for continuous rewards}

In the case of continuous rewards,
the proofs are simpler due to positive $X_{kt}$,
in particular $\lambda = E_g \mu$.
Recall that for continuous rewards,
we assume in condition (B2) that 
\begin{equation} \label{Prho}
\sup_{\mu>0} P_{\mu}(X \leq \gamma \mu) \rightarrow 0 \mbox{ as } \gamma \rightarrow 0.
\end{equation}
Moreover (\ref{E4}) holds and for $0 < \delta \leq 1$,
there exists $\tau_{\delta}>0$ such that for $0 < \theta \mu \leq \tau_{\delta}$,
\begin{eqnarray} \label{tau1}
M_{\mu}(\theta) & \leq & e^{(1+\delta) \theta \mu}, \\ \label{tau2}
M_{\mu}(-\theta) & \leq & e^{-(1-\delta) \theta \mu}. 
\end{eqnarray}
In addition for each $t \geq 1$, 
there exists $\xi_t >0$ such that
\begin{equation} \label{sigbound}
\sup_{\mu \leq \xi_t} P_{\mu}(\wht \sigma_t^2 \leq \gamma \mu^2) \rightarrow 0 \mbox{ as } \gamma \rightarrow 0,
\end{equation}
where $\wht \sigma_t^2 = t^{-1} \sum_{u=1}^t (X_u -\bar X_t)^2$ and $\bar X_t = t^{-1} \sum_{u=1}^t X_u$ for i.i.d. $X_u \stackrel{d}{\sim} F_{\mu}$.

\smallskip
{\sc Proof of Lemma} \ref{lem4}.
To show (\ref{lem4.1}) and (\ref{lem4.2}),
it suffices to show that 
\begin{equation} \label{42.1}
\sup_{\mu \geq d_n} E_{\mu} T_b \leq 1+o(1).  
\end{equation}
Let $\theta > 0$ to be further specified.
By Markov's inequality,
\begin{equation*}
P_{\mu}(S_t \leq b_n t \zeta_n)  \leq  [e^{\theta b_n \zeta_n}M_{\mu}(-\theta)]^t.
\end{equation*}
Moreover, for any $\gamma >0$, 
\begin{equation*}
M_{\mu}(-\theta)  \leq  P_{\mu}(X \leq \gamma \mu)+e^{-\gamma \theta \mu},
\end{equation*} 
hence
\begin{eqnarray} \label{42.3}
E_{\mu} T_b & \leq & 1+\sum_{t=1}^{\infty} P_{\mu}(S_t \leq b_n t \zeta_n) \\ \nonumber
& \leq & \{ 1-e^{\theta b_n \zeta_n}[P_{\mu}(X \leq \gamma \mu)+e^{-\gamma \theta \mu}] \}^{-1}.
\end{eqnarray}
Let $\gamma = \frac{1}{\log n}$ and $\theta = n^{\eta}$ for some $ \omega<\eta <\frac{1}{\beta+1}$.
By (\ref{Prho}), $b_n$ is sub-polynomial in $n$, and $d_n = n^{-\omega}$,
for $\mu \geq d_n$,
$$e^{\theta b_n \zeta_n} \rightarrow 1, 
\quad e^{-\gamma \theta \mu} \rightarrow 0,
\quad P_{\mu}(X \leq \gamma \mu) \rightarrow 0,
$$
and (\ref{42.1}) follows from (\ref{42.3}). $\wbox$

\smallskip
{\sc Proof of Lemma} \ref{lem5}.
By (\ref{E4}), for $\mu$ small,
\begin{eqnarray} \nonumber
\rho_\mu : = E_\mu X^2 & = & E_\mu(X^2 \mathbf{1}_{\lbrace X<1 \rbrace}) + E_\mu(X^2 \mathbf{1}_{\lbrace X \geq 1 \rbrace}) \\ \nonumber
& \leq & E_\mu X + E_\mu X^4 = O(\mu).
\end{eqnarray}
Hence to show (\ref{lem5.1}) and (\ref{lem5.2}),
we proceed as in the proof of Lemma \ref{lem5} for discrete rewards,
applying (\ref{tau2}) in place of (\ref{Mmu2}), 
with any fixed $\theta>0$ in place of $\theta_{\delta}$ in (\ref{case2}) and (\ref{cn3}).
$\wbox$

\smallskip
{\sc Proof of Lemma} \ref{lem6}.
It follows from (\ref{tau1}) with $\theta = \frac{\tau_1}{\mu}$ and Markov's inequality [see (\ref{lem6.2}) and (\ref{com})] that for $n$ large,
\begin{eqnarray*}
	& & P_{\mu}(S_t > tb_n \zeta_n \mbox{ for some } t \geq 1) \\
	& \leq & \sup_{t \geq 1} [e^{-\theta b_n \zeta_n} M_{\mu}(\theta)]^t \leq e^{-\theta(b_n \zeta_n-2 \mu)} \rightarrow 0. \quad \mbox{$\wbox$}
\end{eqnarray*}

\smallskip
{\sc Proof of Lemma} \ref{lem7}.
Let $\eta > 0$ and choose $\delta>0$ such that $(1+\delta)(1-\epsilon) < 1$.
It follows from (\ref{tau1}) with $\theta = \frac{\tau_\delta}{\mu}$ and Markov's inequality that for $u$ large,
\begin{eqnarray} \label{72.2}
& & P_{\mu}(S_t \geq t \zeta_n+c_n \wht \sigma_t \sqrt{t} \mbox{ for some } t > u) \\ \nonumber
& \leq & P_{\mu}(S_t \geq t \zeta_n \mbox{ for some } t > u) \\ \nonumber
& \leq & \sup_{t>u} [e^{-\theta \zeta_n} M_{\mu}(\theta)]^t \leq	e^{-u \theta[\zeta_n-(1+\delta) \mu]} \leq e^{-u \tau_{\delta} [(1-\epsilon)^{-1}-(1+\delta)]} \leq \eta.
\end{eqnarray}

By (\ref{sigbound}),
we can select $\gamma > 0$ such that for $n$ large (so that $\mu  \leq (1-\epsilon)\zeta_n \leq \min_{1 \leq t \leq u} \xi_t)$,
\begin{equation} \label{7eta}
\sum_{t=1}^u P_{\mu}(\wht \sigma_t^2 \leq \gamma \mu^2) \leq \eta.
\end{equation}
Let $\theta = \frac{\tau_1}{\mu}$. By (\ref{tau1}), (\ref{7eta}) and Bonferroni's inequality,
\begin{eqnarray} \label{72.1}
& & P_{\mu}(S_t > t \zeta + c_n \wht \sigma_t \sqrt{t} \mbox{ for some } t \leq u) \\ \nonumber
&\leq & P_{\mu}(S_t \geq  c_n \wht \sigma_t \sqrt{t} \mbox{ for some } t \leq u) \\ \nonumber
& \leq & \eta + \sum_{t=1}^u P_{\mu}(S_t \geq c_n \mu \sqrt{\gamma t}) \\ \nonumber
& \leq & \eta + \sum_{t=1}^u e^{-\theta c_n \mu \sqrt{\gamma t}} M_{\mu}^t(\theta) \\ \nonumber
& \leq & \eta + \sum_{t=1}^u e^{-\tau_1(c_n \sqrt{\gamma t}-2t)} \rightarrow \eta.
\end{eqnarray}
Lemma \ref{lem7} follows from (\ref{72.2}) and (\ref{72.1}) since $\eta$ can be chosen arbitrarily small.
$\wbox$

\section{Proof of Theorem 2} \label{thm3prf}

The idealized algorithm in the beginning of Section 5.1 of the main manuscript captures the essence of how CBT behaves.
We reveal $\mu_k$ when the first positive loss of arm $k$ appears.
If $\mu_k > \zeta$ [with optimality when $\zeta=\zeta_n$, see (5.3) of the main manuscript] then we stop sampling from arm $k$ and sample the next arm $k+1$. 
If $\mu_k \leq \zeta$ then we exploit arm $k$ a further $n$ times before stopping. 

In the idealized version of empirical CBT,
we reveal $\mu_k$ when the first positive loss of arm $k$ appears and stop exploring the arm.
Since the first positive loss of an arm has mean $\lambda$,
the sum of losses after $k$ arms have been played has mean $k \lambda$.
When $\min_{1 \leq i \leq k} \mu_i \leq \wht \zeta_k (:= \tfrac{k \lambda}{n})$ we stop exploring, 
and exploit the best arm a further $n$ times.
More specifically:

\vskip0.5in
\underline{Idealized empirical CBT}

\begin{enumerate}
\item For $k=1,2,\ldots:$ 
Draw $n_k$ rewards from arm $k$,
where
$$n_k = \inf \{ t \geq 1: X_{kt} > 0 \}.
$$

\item Stop when there are $K$ arms,
where
$$K = \inf \Big\{ k \geq 1: \min_{1 \leq i \leq k} \mu_i \leq \tfrac{k \lambda}{n} \Big\}.
$$

\item Draw $n$ additional rewards from arm $j$ satisfying $\mu_j = \min_{1 \leq k \leq K} \mu_k$.
\end{enumerate}

\vskip0.5in
\noi The regret of this algorithm is $R_n' = \lambda EK + n E(\min_{1 \leq k \leq K} \mu_k)$.

\begin{customthm}{2} \label{thmA}
The idealized empirical CBT has regret $R_n' \sim C I_{\beta} n^{\frac{\beta}{\beta+1}}$,
where $C = (\frac{\lambda \beta (\beta+1)}{\alpha})^{\frac{1}{\beta+1}}$ and $I_{\beta} = (\frac{1}{\beta+1})^{\frac{1}{\beta+1}} (2-\frac{1}{(\beta+1)^2}) \Gamma(2-\frac{\beta}{\beta+1})$.
\end{customthm}

{\sc Proof}. 
We stop exploring after $K$ arms,
where
\begin{equation} \label{keq}
K = \inf \{ k: \min_{1 \leq j \leq k} \mu_j \leq \wht \zeta_k \}, \quad \wht \zeta_k = \tfrac{k \lambda}{n}.
\end{equation}
Let
$$D_k^1 = \{ \wht \zeta_k-\tfrac{\lambda}{n} < \min_{1 \leq j \leq k-1} \mu_j \leq \wht \zeta_k \}, 
\quad D_k^2 = \{ \min_{1 \leq j \leq k-1} \mu_j > \wht \zeta_k, \mu_k \leq \wht \zeta_k \}.
$$
We check that $D_k^1$, 
$D_k^2$ are disjoint, 
and that $D_k^1 \cup D_k^2 = \{ K=k \}$.
Essentially $D_k^1$ is the event that $K=k$ and the best arm is not $k$,
and $D_k^2$ the event that $K=k$ and the best arm is $k$.

For any fixed $k \in {\bf Z}^+$, 
\begin{eqnarray} \label{Dk1}
P(D_k^1) & = & [1-p(\wht \zeta_k-\tfrac{\lambda}{n})]^{k-1} - [1-p(\wht \zeta_k)]^{k-1} \\ \nonumber
& = & \{ 1-p(\wht \zeta_k)+[1+o(1)] \tfrac{\lambda}{n} g(\wht \zeta_k) \}^{k-1} - [1-p(\wht \zeta_k)]^{k-1} \\ \nonumber
& \sim & \{ [1-p(\wht \zeta_k)]^{k-1} \} \tfrac{k \lambda}{n} g(\wht \zeta_k) \\ \nonumber
& \sim & \exp(-\tfrac{\alpha \lambda^\beta}{\beta}k^{\beta+1}n^{-\beta}) \alpha \lambda^\beta k^\beta n^{-\beta}.
\end{eqnarray}
Moreover
\begin{equation} \label{RC1}
E(R_n'|D_k^1) \sim k \lambda + n(\tfrac{k \lambda}{n}) = 2 k \lambda.
\end{equation}
Likewise,
\begin{eqnarray} \label{Dk2}
P(D_k^2) & = & \{ [1-p(\wht \zeta_k)]^{k-1} \} p(\wht \zeta_k) \\ \nonumber
& \sim & \exp(-\tfrac{\alpha \lambda^\beta}{\beta}k^{\beta+1}n^{-\beta}) \tfrac{\alpha \lambda^\beta}{\beta} k^\beta n^{-\beta}, \\ \label{RC2}
E(R_n'|D_k^2) & = & k \lambda + n E(\mu|\mu \leq \wht \zeta_k) \\ \nonumber
& = & 2k \lambda -\tfrac{n v(\hat \zeta_k)}{p(\hat \zeta_k)} \sim (2-\tfrac{1}{\beta+1}) k\lambda.
\end{eqnarray}

Combining (\ref{Dk1})--(\ref{RC2}) gives us
\begin{eqnarray} \label{RCp}
R_n' & = & \sum_{k=1}^{\infty} [E(R_C'|D_k^1)P(D_k^1) + E(R_C'|D_k^2)P(D_k^2)] \\ \nonumber
& \sim & \sum_{k=1}^{\infty}  \exp(-\tfrac{\alpha \lambda^{\beta}}{\beta} k^{\beta+1}n^{-\beta})
(\tfrac{\alpha \lambda^{\beta+1}}{\beta} k^{\beta+1}n^{-\beta}) (2 \beta+2-\tfrac{1}{\beta+1}), 
\end{eqnarray}

It follows from (\ref{RCp}) and a change of variables $x=\tfrac{\alpha \lambda^{\beta}}{\beta} k^{\beta+1}n^{-\beta}$ that 
\begin{eqnarray} \nonumber
R_n' & \sim &   (2 \beta+2-\tfrac{1}{\beta+1}) \int_{0}^{\infty} \exp(-\tfrac{\alpha \lambda^{\beta}}{\beta} k^{\beta+1}n^{-\beta})
(\tfrac{\alpha \lambda^{\beta+1}}{\beta} k^{\beta+1}n^{-\beta}) dk \\ \nonumber
& = &  (2 \beta+2-\tfrac{1}{\beta+1}) \int_{0}^{\infty} \tfrac{1}{\beta+1}(\tfrac{\lambda \beta}{\alpha})^{\tfrac{1}{\beta+1}} n^{\tfrac{\beta}{\beta+1}} \exp(-x) x^{\tfrac{1}{\beta+1}} dx\\ \nonumber
& = &  (2 - \tfrac{1}{(\beta+1)^2}) (\tfrac{\lambda \beta}{\alpha})^{\tfrac{1}{\beta+1}} \Gamma(2 - \tfrac{\beta}{\beta+1}) n^{\tfrac{\beta}{\beta+1}},
\end{eqnarray}
and Theorem \ref{thmA} holds.
$\wbox$